%% file: main.tex
\documentclass[journal]{IEEEtran}

\IEEEoverridecommandlockouts                              

\usepackage[ruled]{algorithm2e} 

\SetAlFnt{\small}
\SetAlCapFnt{\small}
\SetAlCapNameFnt{\small}
\SetAlCapHSkip{0pt}
\IncMargin{-\parindent} 

\usepackage{float}
\usepackage{multicol}
\usepackage{multirow}
\usepackage{algpseudocode}
\usepackage{color}
\usepackage{graphicx}
\usepackage{bm}
\usepackage{amsmath}
\usepackage{amssymb}
\usepackage{tabularx}
\usepackage{subcaption}
\usepackage{url}
\usepackage{cite}
\usepackage{prettyref}
\usepackage{booktabs}
\usepackage{siunitx}
\usepackage{makecell}
\usepackage{comment}
\usepackage{wrapfig}
\usepackage{hyperref}

\graphicspath{
    {./fig/}
}

\title{Trajectory Optimization for Collision-Aware Redundant Robotic Multi-Axis Additive Manufacturing by Constrained Gradient Projection}

\author{Zhikai Shen,~\IEEEmembership{Student Member,~IEEE}, Jiasheng Qu, Chenyu Xu, Zhuo Huang, \\ Chengkai Dai, Yongzhe Li, ~\IEEEmembership{Member,~IEEE}, and Guoxin Fang\textsuperscript{\dag},~\IEEEmembership{Member,~IEEE}

\thanks{Z. Shen, J. Qu, C. Xu, Z.Huang and G. Fang are with the Department of Mechanical and Automation Engineering, The Chinese University of Hong Kong, Hong Kong, China. 
C. Dai and G. Fang are with the Centre for Perceptual and Interactive Intelligence (CPII), Hong Kong, China. Y. Li is with the Department of Robotics at
the School of Mechanical Engineering, Southeast University, China. 
}
\thanks{
 \dag~Corresponding Author: \textit{Guoxin Fang (\url{guoxinfang@cuhk.edu.hk})}.}
}

\usepackage{color}
\usepackage[normalem]{ulem}

\definecolor{Rosolic}{cmyk}{0.00,1.00,0.50,0}
\definecolor{bleudefrance}{rgb}{0.19, 0.55, 0.91}
\definecolor{green}{cmyk}{0.3,0.2,0.95,0.0}
\definecolor{brown}{cmyk}{0.4,0.7,1.0,0.5}
\definecolor{dblue}{cmyk}{1,0.97,0.35,0.0}
\definecolor{Cyan}{cmyk}{1,0,0,0}
\definecolor{Azure}{rgb}{0.0, 0.5, 1.0}

\begin{document}

\maketitle

\begin{abstract}
Redundant robotic multi-axis additive manufacturing (MAAM) enables support-free and conformal fabrication, but trajectory optimization for long-horizon path remains challenging under strict deposition-position constraints and time-varying collision constraints.
This work propose a computational framework for collision-aware trajectory optimization in redundant robotic MAAM. We first formulate nozzle-workpiece relative kinematics using a relative Jacobian, and develop a differentiable SDF-based collision model that captures fabrication-induced geometry evolution and provides optimization gradients. The deposition position is then enforced as a hard waypoint-wise equality constraint through iterative projection onto the self-motion manifold, with the loss gradient restricted to the corresponding tangent space. 
Experiments on an 8-DOF robotic MAAM platform with diverse long-horizon support-free and conformal toolpaths show that our method maintains a mean nozzle-position error below $10~\mu\mathrm{m}$, reduces maximum joint jerk by up to $77.6\%$, and eliminates all sampled collision and orientation violations. Compared with the SQP-based baseline, it achieves up to a $10.2\times$ speedup and improved convergence. Physical fabrication experiments further verify that the resulting smooth, collision-free trajectories enable successful printing of complex geometries with fewer visible deposition artifacts.
\end{abstract}



\input{tex/intro.tex}
\input{tex/formulation.tex}
\input{tex/optimization.tex}
\input{tex/experiment.tex}
\input{tex/conclusion.tex}
\input{tex/appendix.tex}

\bibliographystyle{IEEEtran}
\bibliography{TMECHTO}

\end{document}

%% file: tex/intro.tex
\section{Introduction}
\label{sec:Introduction}

\IEEEPARstart{M}{ulti}-axis additive manufacturing (MAAM) enables spatial material alignment through local control of the nozzle's position and orientation, achieving advanced functionalities such as support-structure elimination~\cite{dai2018support,wu2017robofdm}, improved surface quality~\cite{sun2026robot}, and tailored anisotropic strength of the fabricated model~\cite{fang2020reinforced, fang2024exceptional}. Redundant robotic systems are widely used to provide the motion freedom and workspace required by MAAM~\cite{zhang2021singularity,yang2025smooth}. In these systems, the available degrees of freedom (DOFs) exceed those required by the deposition task, providing extra freedom that can be exploited for motion-related objectives~\cite{marcotte2025robotic, rescsanski2026constrained}. 

\begin{figure*}[t]
    \centering
\includegraphics[width=0.9\linewidth]{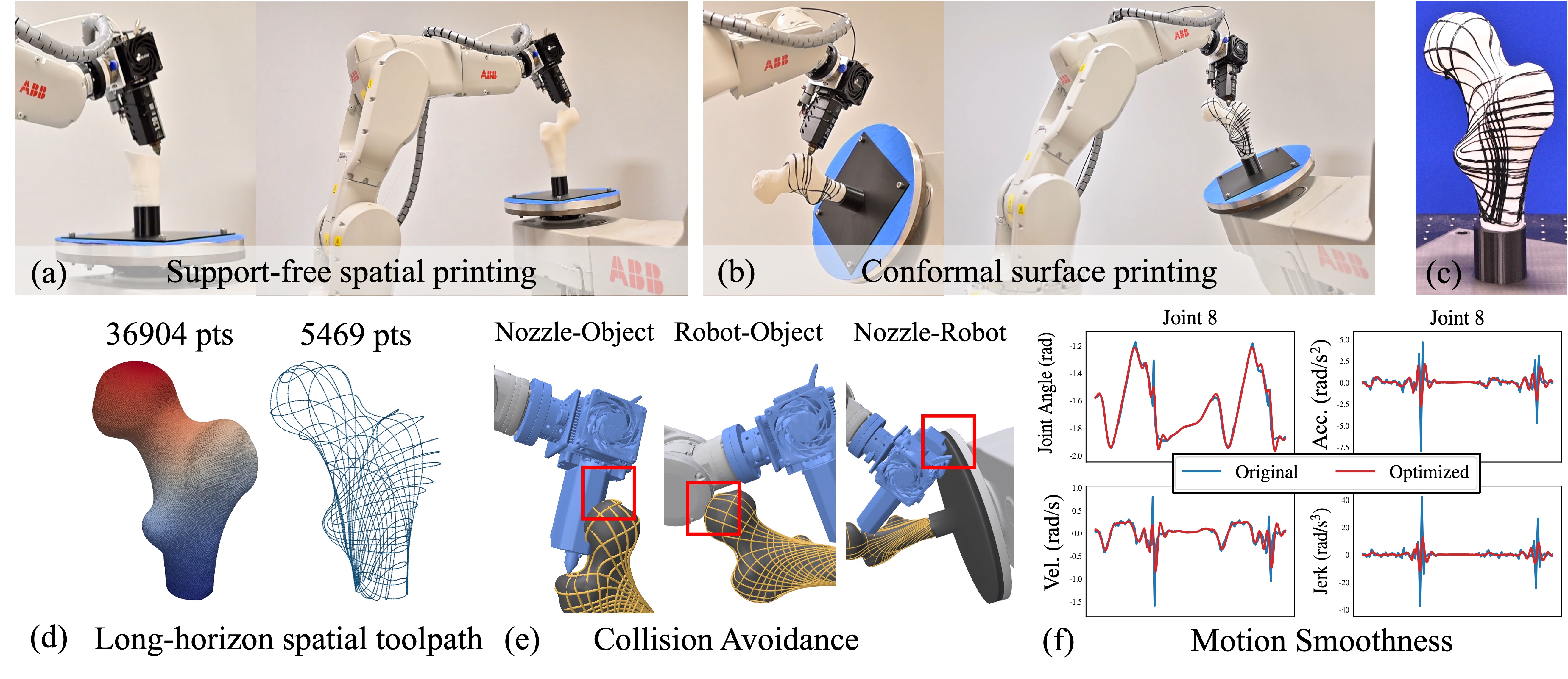}
\caption{
Overview of redundant robotic MAAM scenarios and trajectory optimization challenges. 
An 8-DOF platform provides redundant motion for complex freeform printing. Two representative scenarios are considered: (a) support-free spatial printing, which builds freeform structures from a substrate, and (b) conformal surface printing, which deposits material on an existing object. (c) The resulting as-printed part.
The motion planning for the fabrication task requires (d) handling long-horizon toolpaths with many waypoints with good scalability, (e) satisfying collision-free constraints, and (f) generating optimized smooth joint trajectories with reduced acceleration/jerk spikes.}
    \label{fig:Teaser}
\end{figure*}

An example of the redundant robotic MAAM system is illustrated in Fig.~\ref{fig:Teaser}. The setup has 8 DOFs in total, combining an ABB IRB 1200 robot arm (6 DOFs) with an ABB IRBP A 250 positioner (2 DOFs). Two representative multi-axis AM processes can be successfully conducted with this setup: 1) support-free spatial printing~\cite{qu2025inf}, where a freeform geometry is printed without support structures through the synchronized motion of the setup~\ref{fig:Teaser}(a); and 2) conformal surface printing~\cite{zhang2023robot}, which deposits material onto existing objects by locally adjusting the nozzle orientation~\ref{fig:Teaser}(b). 
For both tasks, the redundant motion of the robotic setup admits infinitely many joint configurations that follow the same toolpath, opening a large solution space for optimizing motion- and fabrication-related objectives (e.g., motion smoothness and collision avoidance)~\cite{liao2021optimization, bai2026tool}.

\subsection{Challenges in Trajectory Optimization and Related Work}
\label{Related Work}

Although redundant robotic MAAM provides the motion freedom needed for complex printing tasks, exploiting this freedom for efficient trajectory optimization over long-horizon toolpaths remains challenging. The main complexities arise from three aspects as follows:

\begin{itemize}
    \item \textbf{High-precision path following with smooth motion.} 
    For the MAAM process, the nozzle must accurately follow the prescribed deposition path while satisfying printing-orientation requirements. 
    This path tracking must be achieved together with smooth redundant motion~\ref{fig:Teaser}(f), as abrupt motion changes and high jerk can degrade deposition stability and printing quality~\cite{dai2020planning, chen2025co}.

    \item \textbf{Evolving collision constraints.} 
    During printing the deposited material continuously changes the workpiece geometry and becomes a dynamically evolving obstacle to subsequent motion. 
    The optimizer must handle robot-workpiece and robot-robot collisions along the trajectory (Fig.~\ref{fig:Teaser}(e)), where the collision constraints are non-convex and expensive to evaluate~\cite{lu2021collision, weingartshofer2023optimization, wen2022path}.

    \item \textbf{Scalability of the optimization.} 
    MAAM toolpaths often contain thousands to tens of thousands of waypoints (see Fig.~\ref{fig:Teaser}(d)). Directly solving the full constrained problem is computationally expensive, whereas waypoint-wise optimization can produce discontinuous motions~\cite{schulman2014motion, kalakrishnan2011stomp}.

\end{itemize}

These challenges call for a unified optimization framework that preserves deposition accuracy, handles evolving collision constraints, and resolves redundancy efficiently over long-horizon toolpaths with good scalability. 
However, existing methods often address only part of this problem and fail to provide a sufficient solution for MAAM with a redundant robotic setup.

For the task of path-constrained planning with redundant robotic systems, a classical approach is the family of penalty-based methods, including STOMP~\cite{kalakrishnan2011stomp}, CHOMP~\cite{ratliff2009chomp}, and path-wise inverse kinematics (IK) solver~\cite{kang2020torm}. They are computationally efficient and have been widely used to optimize smoothness while considering reconfiguration costs~\cite{weingartshofer2023optimization, lu2021collision, wang2023fast}. However, because the deposition requirements are applied only as soft penalties, position accuracy is inherently traded off against smoothness via the penalty weights. 
Exact path enforcement can instead be achieved by parameterizing task-space redundancy, such as the free rotation about the nozzle axis~\cite{chen2026gradient, beschi2019optimal}. While this preserves tracking accuracy, it relies on an explicit low-dimensional parameterization~\cite{burdick1989inverse} and does not scale to high-redundancy systems (e.g., the one shown in Fig.~\ref{fig:Teaser}), where redundancy cannot be captured by a single intuitive parameter. On the other hand, sequential quadratic programming (SQP)-type methods\cite{schulman2014motion, bonalli2019trajectory} enforce constraints explicitly, and have been applied to jerk-aware trajectory optimization~\cite{dai2020planning,chen2025co, chen2026gradient}. By imposing the constraints directly, they preserve path-following accuracy while resolving redundancy within the optimization as additional decision variables, which naturally accommodates high-dimensional redundant systems~\cite{chen2025co}. However, SQP-based methods generally incur high computational costs for long-horizon trajectory optimization tasks, since each iteration solves a quadratic subproblem whose size grows with the number of variables and constraints, leading to sharply increasing per-iteration cost with problem dimension~\cite{betts2010practical}.

For most of these works, collision handling is another critical component that remains insufficiently addressed. Many trajectory optimizers rely on precomputed or static collision models~\cite{schulman2014motion, ratliff2009chomp}, which assume a known and unchanging environment throughout planning~\cite{lu2021collision, weingartshofer2023optimization, wen2022path}. As discussed above, for additive manufacturing tasks, simplifying the collision model as static can greatly restrict the feasible search space~\cite{qu2025inf}. To cope with such constraints, prior works typically rely on a graph-search stage that first discretizes the redundant degrees of freedom and prunes colliding configurations, after which a continuous optimizer refines the surviving candidates~\cite{dai2020planning, zhang2021singularity}. While this can be effective on short toolpaths (i.e., with hundreds of waypoints~\cite{chen2025co}), the search graph grows combinatorially with the number of waypoints and the redundancy dimension. There is no existing work that simultaneously addresses long-horizon MAAM trajectory optimization with dynamically evolving collision constraints.

\subsection{Our Method and Contribution}
\label{subsec:Method and Contribution}
In this paper, we propose a scalable gradient-projection-based trajectory optimization framework for redundant robotic MAAM that enforces deposition-position accuracy through manifold projection and incorporates evolving collision constraints through a differentiable SDF-based penalty model. The key idea is to treat the path-following requirement as a per-waypoint equality constraint defining a self-motion manifold, and to project each gradient update onto its tangent space. Operating directly in joint space, this projection requires no explicit low-dimensional parameterization of the redundancy and thus naturally accommodates high-dimensional redundant systems. Unlike conventional penalty-based methods, which trade deposition accuracy for smoothness via tuned weights, or SQP-type methods, which repeatedly solve costly constrained subproblems, the proposed projection maintains deposition-path accuracy to numerical precision while retaining the low per-iteration cost of gradient-based optimization, making it scalable to long-horizon toolpaths.

To handle the continuously evolving workpiece, we propose a fabrication-specific differentiable collision-checking model based on an implicit signed distance field (SDF) representation. At each trajectory point, collision checking is performed between the already deposited portion of the toolpath and the robot configuration. This models the incrementally growing workpiece while providing differentiable collision information for gradient-based redundancy resolution, avoiding a separate graph-search pre-filtering stage. Since the SDF queries and their gradients are independent across waypoints, the collision model integrates naturally with the projected-gradient optimizer and can be evaluated in parallel on the GPU for efficient optimization.

The technical contributions of this paper are summarized as follows:

\begin{itemize}
    \item We formulate redundant robotic MAAM trajectory planning as a long-horizon hard path-constrained optimization problem that jointly optimize for smoothness, orientation consistency, dynamic feasibility, and collision avoidance.

    \item We develop an differentiable collision model tailored to MAAM, which represents the accumulated printed geometry along the deposition sequence and provides collision gradients for trajectory optimization.

    \item We propose a trajectory-level gradient projection optimizer that alternates between self-motion manifold projection and tangent-space redundancy optimization, preserving tracking accuracy without explicit redundancy parameterization.
\end{itemize}

The proposed framework is evaluated on both support-free MAAM of free-form models and conformal printing on complex surfaces. Across six tested cases, it maintains a mean waypoint-position error below $0.01$ mm (0.005\% of the model size), obtains collision-free and orientation-consistent trajectories under the adopted sampled SDF collision model, and achieves up to an order-of-magnitude (up to 10.2 times) speedup compared with the SQP-based baseline and a 77.6\% reduction in maximum jerk. 

%% file: tex/formulation.tex
\section{Trajectory Optimization for Collision-Free Multi-Axis Additive Manufacturing}
\label{sec:Problem Formulation}

In this section, we first give the formulation of the trajectory optimization problem for MAAM, introduce the relative nozzle--object kinematics for a general redundant robotic setup, and propose the SDF-based differentiable collision model that accounts for the accumulated printed geometry.

\subsection{Problem Formulation for MAAM Trajectory Optimization}
\label{subsec:TO Formulation}

Consider a MAAM task performed to fabricate object $\mathcal{O}$ by a robotic system with \(n\) DOFs. 
The task requires driving the end-effector along a predefined spatial printing path \(\mathcal{P} = \{\mathbf{p}_i\}_{i=1}^{N}\), with each point \(\mathbf{p}_i\) assigned a reference support-free printing direction \(\mathbf{n}_i\). Both \(\mathbf{p}_i\) and \(\mathbf{n}_i\) are generated by a field-based slicer~\cite{qu2025inf} to ensure support-free printability (Fig.~\ref{fig:setup_diagram}(a)). For conformal surface printing, the path follows principal stress-guided conformal toolpaths~\cite{wang2026topology} to strengthen the model, while \(\mathbf{n}_i\) is defined as the local surface normal to maintain perpendicular deposition to the surface (see Fig.~\ref{fig:Teaser}(c), Fig.~\ref{fig:socket_result}, Fig.~\ref{fig:dome}). The precomputed paths and orientations are used as input to our motion planning framework.

The trajectory optimization problem seeks a sequence of feasible joint configurations \(\mathcal{Q} = \{\mathbf{q}_i\}_{i = 1, \ldots, N}\), where \(\mathbf{q}_i\in\mathbb{R}^n\), that minimizes dynamic costs while satisfying kinematic and safety requirements:
\begin{align} 
    \min_{\mathcal{Q}} 
    & \ \ \ \ \sum_{i=1}^N\Phi_i(\mathbf{v}_i,\mathbf{a}_i,\mathbf{j}_i) \label{eq:motionSmoothness}\\
    \text{s.t.} 
    & \ \ \ \ \mathbf{FK}_{\mathbf{p}}(\mathbf{q}_i) = \mathbf{p}_i,
    \label{eq:pathConstraint} \\ 
    & \ \ \ \ \left\langle \mathbf{FK}_{\mathbf{n}}(\mathbf{q}_i),\mathbf{n}_i \right\rangle 
    \geq \cos\theta_{\max},
    \label{eq:oriConstraint}\\
    & \ \ \ \ \left\langle \mathbf{FK}_{\mathbf{n}}^{m}(\mathbf{q}_i),\mathbf{g} \right\rangle 
    \geq \cos\theta_{\max}^{g},
    \label{eq:gravityOriConstraint}\\
    & \ \ \ \ \Gamma(\mathcal{O}_i, \mathbf{q}_i) = 0,
    \label{eq:colConstraint} \\ 
    & \ \ \ \ \mathbf{q}_{min} \leq \mathbf{q}_i \leq \mathbf{q}_{max},
    \label{eq:jointLimits} \\ 
    & \ \ \ \ \vert{\mathbf{v}_i}\rvert \leq \mathbf{v}_{max},\ 
    \vert{\mathbf{a}_i}\rvert \leq \mathbf{a}_{max},\ 
    \vert{\mathbf{j}_i}\rvert \leq \mathbf{j}_{max}.
    \label{eq:dynLimits}
\end{align}

Here, the objective \(\Phi_i\) penalizes velocity, acceleration, and jerk to ensure motion smoothness:
\begin{equation}
    \Phi_i
    =
    \omega_{\mathbf{v}} \mathbf{v}_i^{\mathrm{T}} \mathbf{W}\mathbf{v}_i
    +
    \omega_{\mathbf{a}} \mathbf{a}_i^{\mathrm{T}}\mathbf{W}\mathbf{a}_i
    +
    \omega_{\mathbf{j}} \mathbf{j}_{i}^{\mathrm{T}}\mathbf{W}\mathbf{j}_{i}.
    \label{eq:DynMetric}
\end{equation}
\(\omega_{\mathbf{v}}\), \(\omega_{\mathbf{a}}\), and \(\omega_{\mathbf{j}}\) are weighting coefficients for the velocity, acceleration, and jerk penalties, respectively, while \(\mathbf{W}\) is the diagonal joint weight matrix. 
The time interval \(\Delta t_i\) is precomputed before optimization, typically as 
\(\Delta t_i = \|\mathbf{p}_{i+1}-\mathbf{p}_i\|/v_{\mathrm{print}}\), where \(v_{\mathrm{print}}\) is the prescribed constant printing speed of the end-effector. 

Constraints Eq.~\eqref{eq:pathConstraint} and Eq.~\eqref{eq:oriConstraint} enforce the toolpath requirements: the end-effector must reach position \(\mathbf{p}_i\) while maintaining the tool orientation error within a tolerance \(\theta_{\max}\) (selected as $30^\circ$ for high surface finishing~\cite{zhang2021singularity}). 
Eq.~\eqref{eq:gravityOriConstraint} further requires the nozzle direction of the main printing arm to remain within \(\theta_{\max}^{g}\) of the gravity direction \(\mathbf{g}\), which helps maintain a feasible deposition posture. 
Additionally, Eq.~\eqref{eq:colConstraint} denotes the collision-free condition, which is later handled through a differentiable SDF-based penalty during optimization. Finally, Eq.~\eqref{eq:jointLimits} and Eq.~\eqref{eq:dynLimits} specify the joint-limit and dynamic-feasibility requirements.

\subsection{Relative Kinematics}
\label{subsec:Relative Kinematics}

\begin{figure}[t]
    \centering
    \includegraphics[width=0.95\linewidth]{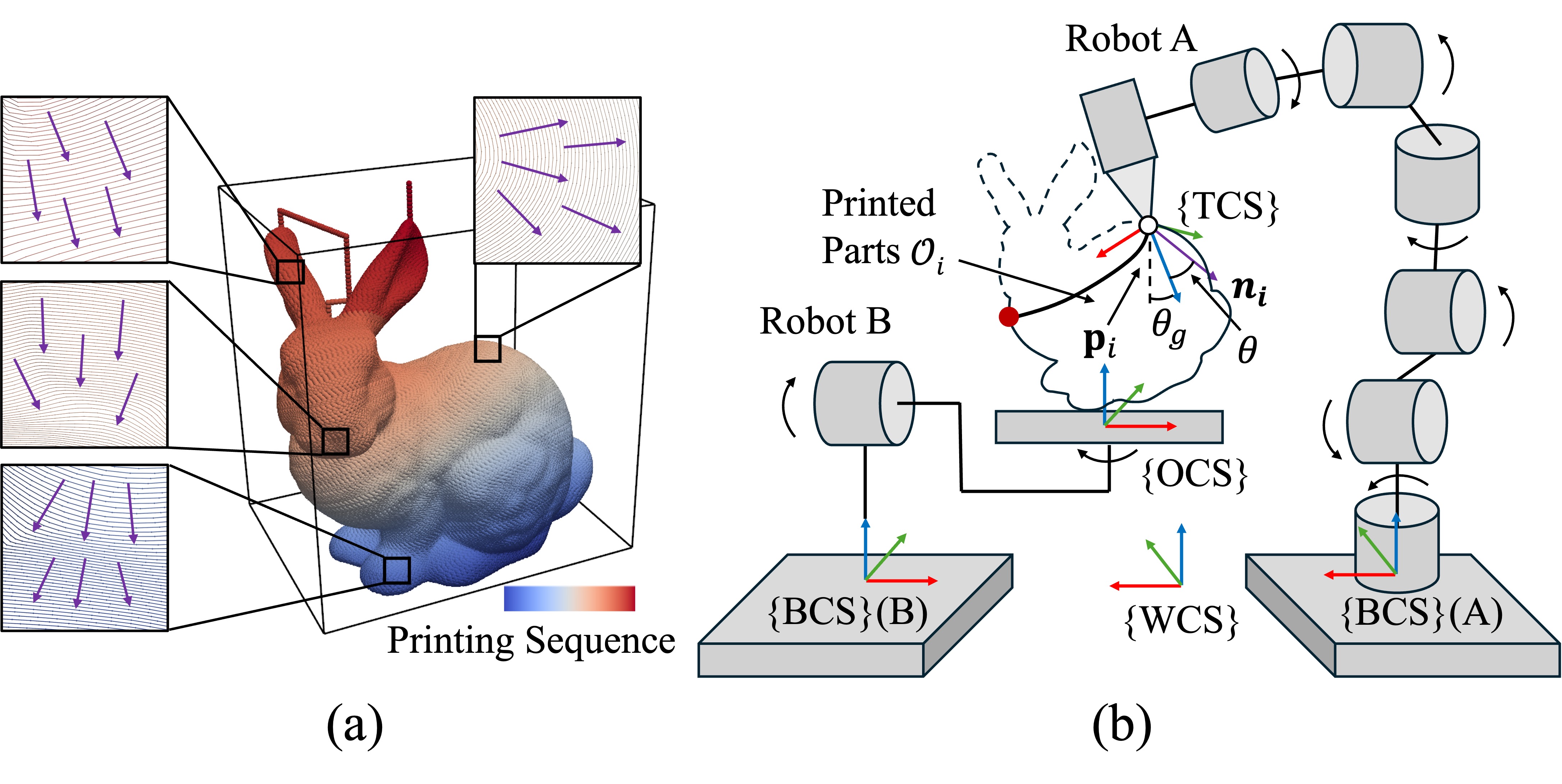}
    \caption{Robotic MAAM setup and task definition. 
    (a) A representative long-horizon toolpath on a Stanford-bunny shell. Each waypoint specifies a deposition point $\mathbf{p}_i$ and a reference support-free printing direction $\mathbf{n}_i$ (purple vectors), with the color map indicating the printing sequence. 
    (b) Dual-robot setup, in which Robot A carries the nozzle while Robot B holds the object, providing higher effective DOFs to enable printing at challenging points, e.g., the red point.
    }
    \label{fig:setup_diagram}
\end{figure}

For the motion planning with dual-robot setup, the path and orientation constraints in Eqs.~\eqref{eq:pathConstraint}--\eqref{eq:gravityOriConstraint} are defined by the pose of the nozzle relative to the object rather than by the absolute pose of either robot alone. Therefore, we derive the nozzle–object relative kinematics for redundant robotic MAAM, which provides the constraint functions and their gradients for trajectory optimization.

As illustrated in Fig.~\ref{fig:setup_diagram}(b), Robot A carries the nozzle while Robot B holds and reorients the object. The deposition task is defined by the relative pose between object coordinate system $\{\mathrm{OCS}\}$ and tool coordinate system $\{\mathrm{TCS}\}$, 
described by 
a single differential-kinematics relation that express the twist of \{TCS\} relative to \{OCS\} as
\begin{equation}
    {}^{O}\boldsymbol{\xi}_{rel}
    =
    \mathbf{J}_{rel}(\mathbf{q}_A,\mathbf{q}_B)
    \begin{bmatrix}
        \mathbf{v}_A \\
        \mathbf{v}_B
    \end{bmatrix}.
    \label{eq:RelativeVelocity}
\end{equation}
Here \({}^{O}(\cdot)\) denotes a quantity expressed in \(\{\mathrm{OCS}\}\), and the target Cartesian twist is written as 
\({}^{O}\boldsymbol{\xi}_{rel}\). \(\mathbf{q}_A, \mathbf{q}_B, \mathbf{v}_A, \mathbf{v}_B\) denote the joint angles and velocities of Robot A and Robot B, respectively. 
\(\mathbf{J}_{rel}(\mathbf{q}_A,\mathbf{q}_B)\) is the relative Jacobian that maps 
the stacked joint velocities of both robots to \({}^{O}\boldsymbol{\xi}_{rel}\). 

\begin{figure}[t]
    \centering
\includegraphics[width=0.5\linewidth]{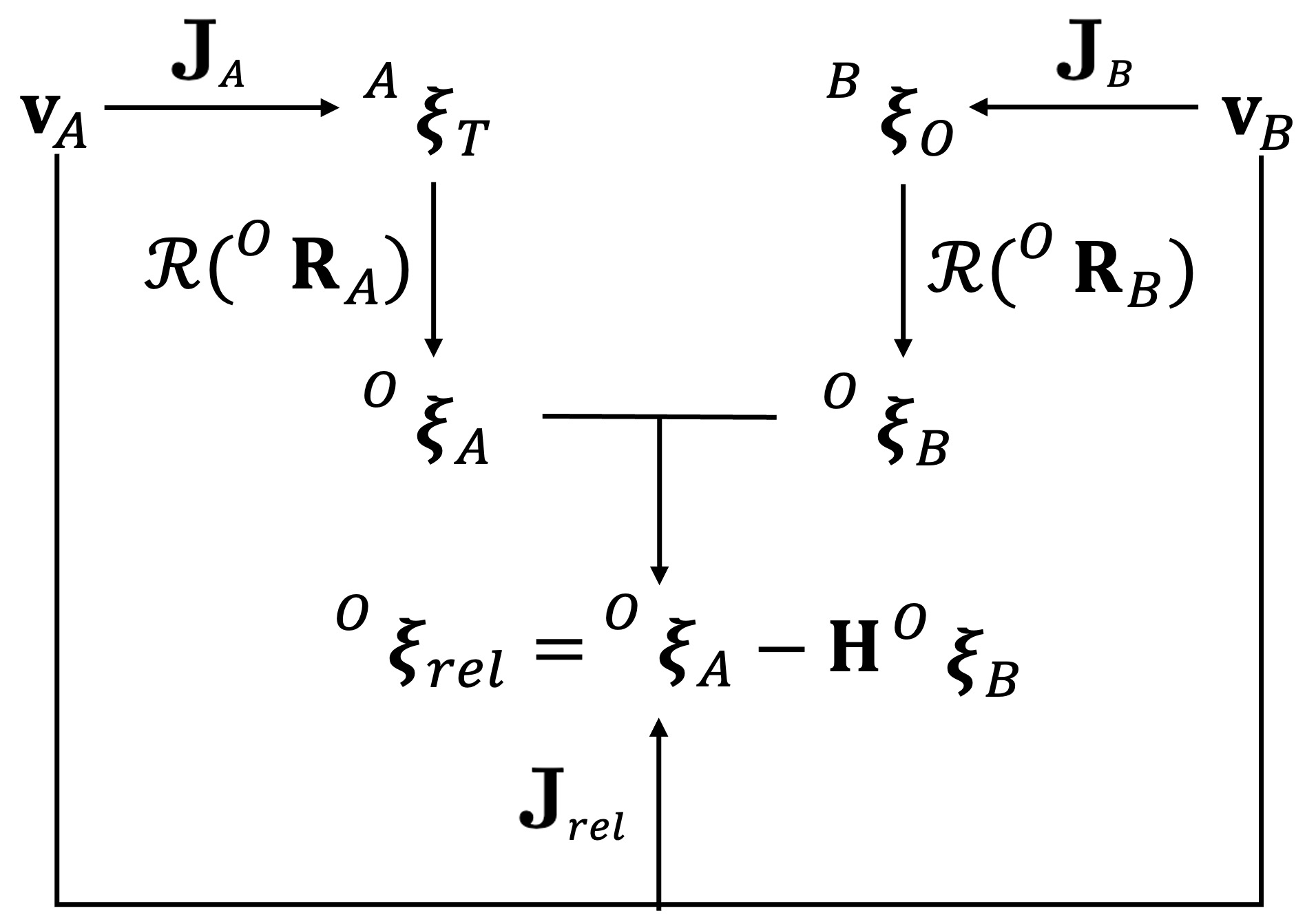}
    \caption{
Construction of the relative Jacobian \(\mathbf{J}_{rel}\), mapping the joint 
velocities of both robots to the relative twist \({}^{O}\boldsymbol{\xi}_{rel}\). 
}
    \label{fig:relative_kinematics_flow}
\end{figure}

According to the relative kinematics (summarized in Fig.~\ref{fig:relative_kinematics_flow}),
\begin{equation}
\begin{aligned}
        &{}^{O}\boldsymbol{\xi}_{rel}=
        \begin{bmatrix}
            {}^{O}\dot{\mathbf{p}}_{T}-{}^{O}\dot{\mathbf{p}}_{O}-{}^{O}\mathbf{r}_{OT}\times{}^O\boldsymbol{\omega}_O \\
            {}^O\boldsymbol{\omega}_T-{}^O\boldsymbol{\omega}_O
        \end{bmatrix} =
    {}^{O}\boldsymbol{\xi}_T
    -
    \mathbf{H}{}^{O}\boldsymbol{\xi}_O \\
    &=
    \underbrace{
    \begin{bmatrix}
        \mathcal{R}({}^{O}\mathbf{R}_{A})\mathbf{J}_A(\mathbf{q}_A)
        &
        -\mathbf{H}
        \mathcal{R}({}^{O}\mathbf{R}_{B})\mathbf{J}_B(\mathbf{q}_B)
    \end{bmatrix}
    }_{\mathbf{J}_{rel}(\mathbf{q}_A,\mathbf{q}_B)}
    \begin{bmatrix}
        \mathbf{v}_A \\
        \mathbf{v}_B
    \end{bmatrix}.
\end{aligned}
\end{equation}
Here, the twist $\boldsymbol{\xi}$ is written in the stacked form of linear and angular velocity, i.e.,
$
\boldsymbol{\xi}
=
\begin{bmatrix}
    \dot{\mathbf{p}}^\top & \boldsymbol{\omega}^\top
\end{bmatrix}^\top .
$
$\boldsymbol{\xi}_T$ and $\boldsymbol{\xi}_O$ denote the twists of \{TCS\} and \{OCS\}, respectively. The shift matrix
\begin{equation}
\mathbf{H}
=
\begin{bmatrix}
    \mathbf{I} & -\mathbf{S}({}^{O}\mathbf{r}_{OT}) \\
    \mathbf{0} & \mathbf{I}
\end{bmatrix}
\end{equation}
accounts for the effect of $\boldsymbol{\xi}_{T}$ and $\boldsymbol{\xi}_{O}$ on $\boldsymbol{\xi}_{rel}$. $\mathbf{S}(\cdot)$ denotes the skew-symmetric matrix, and ${}^{O}\mathbf{r}_{OT}$ is the vector from the origin of \{OCS\} to the origin of \{TCS\}. $\mathbf{J}_A$ and $\mathbf{J}_B$ are Jacobians of corresponding robot, and ${}^{O}\mathbf{R}_{A}, {}^{O}\mathbf{R}_{B}\in SO(3)$ are rotations from the base coordinate system (remark as \{BCS\}) of corresponding robot to \{OCS\}. The block-diagonal operator
$\mathcal{R}(\mathbf{R}) = diag(\mathbf{R},\mathbf{R})$ applies the rotation $\mathbf{R}$ to both the linear and angular components of the twist.

With the defined relative Jacobian, the differentiated form of the tool-orientation constraint (i.e., Eq.~\eqref{eq:oriConstraint}) can be computed for subsequent gradient-based optimization (see detail in Sec.~\ref{subsec:constrained_gradient_projection}). Specifically, we reformulate the constraint as

\begin{equation}
    g_{ori}(\mathbf{q}_i)
    =
    \cos\theta_{\max}
    -
    \left\langle
    \mathbf{FK}_{\mathbf{n}}(\mathbf{q}_i),
    \mathbf{n}_i
    \right\rangle
    \leq 0 ,  \label{eq:oriConstraintScalar}
\end{equation}
and its gradient w.r.t. each robot joint can be expressed by
\begin{equation}
\begin{aligned}
        \nabla_{\mathbf{q}_i}&g_{ori}(\mathbf{q}_i)
        =-\left\langle\frac{\partial \mathbf{FK}_{\mathbf{n}}(\mathbf{q}_i)}{\partial \mathbf{q}_i},\mathbf{n}_i\right\rangle 
        = -\left\langle\frac{\partial {}^{O}\mathbf{R}_{T}(\mathbf{q}_i)}{\partial \mathbf{q}_i}\mathbf{e}_{\mathbf{n}},\mathbf{n}_i\right\rangle \\
        &= \left[ \mathbf{S}({}^{O}\mathbf{R}_{T}\mathbf{e}_{\mathbf{n}})\frac{\partial\boldsymbol{\omega}}{\partial\mathbf{q}_i} \right ]^{\top}\mathbf{n}_i
        =-\mathbf{J}_{rel,\boldsymbol{\omega}}^\top\mathbf{S}({}^{O}\mathbf{R}_{T}\mathbf{e}_{\mathbf{n}})\mathbf{n}_i ,
\end{aligned}
\end{equation}
where \(\mathbf{e}_{\mathbf{n}}\in\mathbb{R}^{3}\) is the fixed unit 
nozzle direction in \(\{\mathrm{TCS}\}\) (e.g., \(\mathbf{e}_{\mathbf{n}} = [0,0,1]^\top\) 
in most cases) and \(\mathbf{J}_{rel, \boldsymbol{\omega}}\) 
is the angular part of $\mathbf{J}_{rel}$. The gravity-related constraint in Eq.~\eqref{eq:gravityOriConstraint} can be handled and differentiated in a similar form by defining
\begin{equation}
    g_{grav}(\mathbf{q}_i)
    =
    \cos\theta^g_{\max}
    -
    \left\langle
    \mathbf{FK}^m_{\mathbf{n}}(\mathbf{q}_i),
    \mathbf{g}
    \right\rangle
    \leq 0,
\end{equation}
where the gradient computed as \begin{equation} \nabla_{\mathbf{q}_i}g_{grav}(\mathbf{q}_i)
= -\mathbf{J}_{A,\boldsymbol{\omega}}^\top\mathbf{S}({}^{A}\mathbf{R}_{T}\mathbf{e}_{\mathbf{n}})\mathbf{g}.
\end{equation}

\subsection{Differentiable Collision Model for MAAM}
\label{subsec:Modeling of Collision}

As mentioned before, effectively handling the collision-free constraint in Eq.~\eqref{eq:colConstraint} is another challenge in trajectory optimization. The difficulty comes from the time-varying nature of the collision geometry. In MAAM process, the robot geometry changes with the configuration $\mathbf{q}$, while the printed object continuously grows along the deposition path. Rebuilding a global collision representation at each iteration is therefore inefficient and undesirable for gradient-based optimization. 

\begin{figure}[t]
    \centering
    \includegraphics[width = 0.9\linewidth]{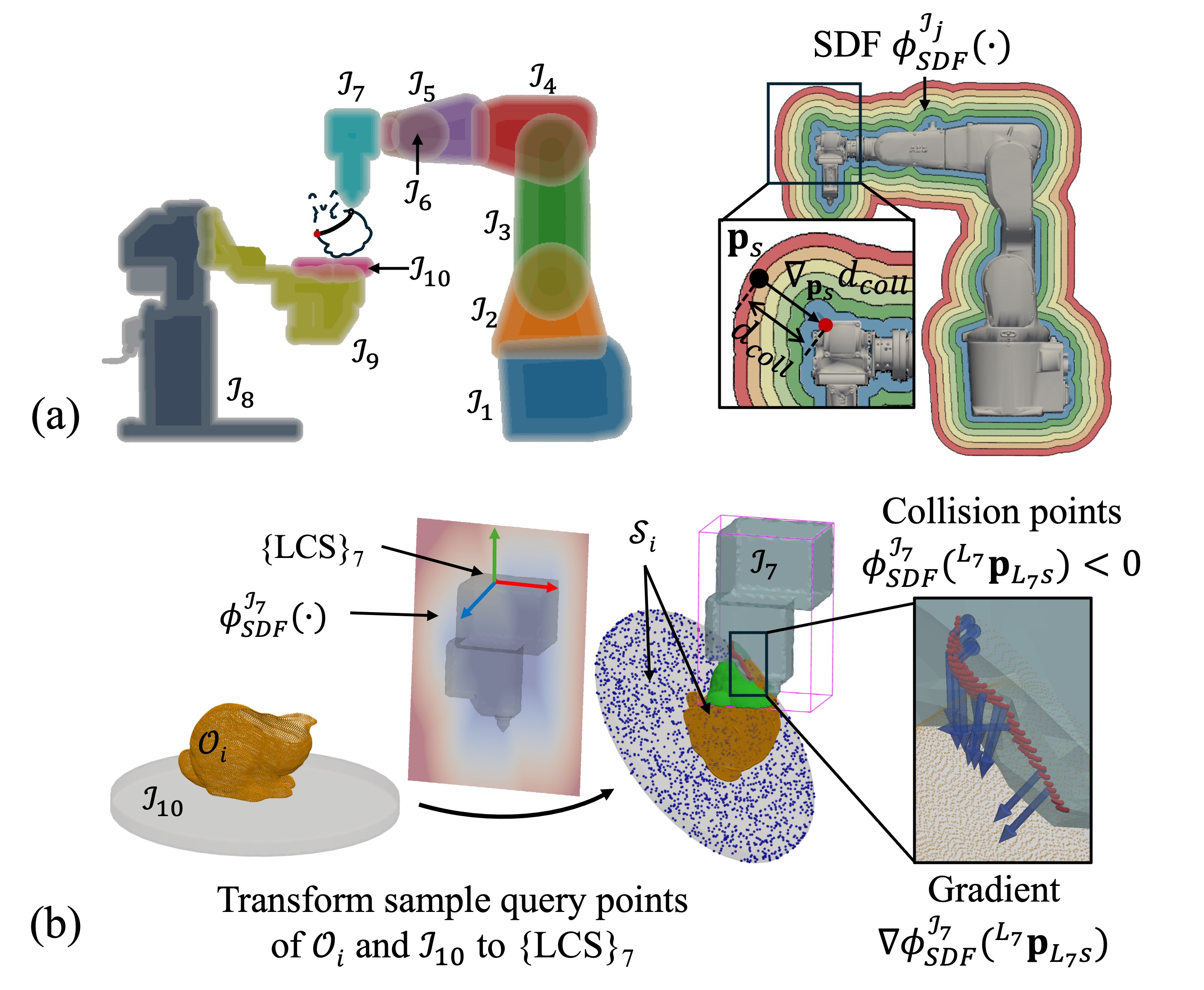}

\caption{
Illustration of the proposed differentiable collision modeling and collision-avoidance formulation.
(a) Differentiable collision modeling using link-wise SDF. 
For each robot link, an SDF $\phi_{\mathrm{SDF}}$ is constructed to evaluate the collision distance $d_{\mathrm{coll}}$ and its spatial gradient $\nabla_{\mathbf{p}_s} d_{\mathrm{coll}}$ at sampled query points. 
(b) An example of collision checking for link $\mathcal{I}_7$ (with the $\mathcal{O}_i$ and $\mathcal{I}_{10}$) is shown. 
Sampled query points $\mathbf{p}_s\in\mathcal{S}_i$ are transformed into the local coordinate frame of $\mathcal{I}_7$ (i.e., \{LCS\}$_7$). 
A predefined link bounding box $\mathcal{B}_{L_7}$ filters the query points, shown as green points. 
Collision points are highlighted in red, and the corresponding SDF gradients indicate the local direction of increasing signed distance.}
\label{fig:collision_modeling}
\end{figure}

To address this issue, we propose a differentiable collision model that decouples the time-varying printed object from the fixed robot-link geometries. The key idea is to precompute an individual SDF for each rigid robot link and to represent the growing printed object using sampled query points. Specifically, for each robot's link and printing nozzle $\mathcal{I}_j \in \{\mathcal{I}\}$, the individual SDF $\phi^{\mathcal{I}_j}_{\mathrm{SDF}}(\cdot)$ is precomputed in its local coordinate frame $\{\text{LCS}\}_j$. As illustrated in Fig.~\ref{fig:collision_modeling}(a), given a query point $\mathbf{p}_s$, the SDF returns its signed distance to the surface of link $\mathcal{I}_j$, while its spatial gradient $\nabla \phi^{\mathcal{I}_j}_{\mathrm{SDF}}(\cdot)$ provides the local direction of increasing signed distance. This individual-SDF representation avoids reconstructing a global SDF for time-varying obstacles and preserves gradient compatibility during trajectory optimization.

On the other hand, the growing printed object is
represented by sampled query points from the deposited material blocks at
each waypoint. Let \(\mathcal{S}_i\) denote the set of query points approximating the current geometry of all obstacles at step \(i\), for each query point \(\mathbf{p}_s \in \mathcal{S}_i\) at configuration $\mathbf{q}_i$, we first transforming the geometry into \{LCS\}$_j$ (see Fig.~\ref{fig:collision_modeling}(b)), the collision distance and the corresponding spatial gradient with respect to link \(\mathcal{I}_j\) can then be queried effectively using pre-constructed SDF as
\begin{equation}
\begin{aligned}
        d_{coll}^{\mathcal{I}_j}(\mathbf{p}_s, \mathbf{q}_i)
    &=
    \phi_{SDF}^{\mathcal{I}j}({}^{L_j}\mathbf{p}_{L_js}), \\
    \ 
    \nabla_{{}^{L_j}\mathbf{p}_{L_js}} d_{coll}^{\mathcal{I}_j}(\mathbf{p}_s, \mathbf{q}_i)
    &=
    \nabla \phi^{\mathcal{I}_j}_{SDF}({}^{L_j}\mathbf{p}_{L_js}).
\end{aligned}
    \label{eq:collision_distance}
\end{equation}
Here, \(\phi_{SDF}^{\mathcal{I}_j}(\cdot)\) and \(\nabla\phi_{SDF}^{\mathcal{I}_j}(\cdot)\) denote the precomputed SDF and spatial gradient of link \(\mathcal{I}_j\), and ${}^{L_j}\mathbf{p}_{L_js}$ denotes the vector from the origin of \{LCS\}$_j$ to $\mathbf{p}_s$, expressed in \{LCS\}$_j$. Using Eq.~\eqref{eq:collision_distance}, the collision constraint in Eq.~\eqref{eq:colConstraint} can be converted into differentiable functions over $\mathbf{q}_i$,
\begin{equation}
    g^{\mathcal{I}_j,\mathbf{p}_s}_{coll}(\mathbf{q}_i) =- d_{coll}^{\mathcal{I}_j}(\mathbf{p}_s,\mathbf{q}_i)\leq0, \ \forall \mathcal{I}_j\in\mathcal{I}, \ \forall\mathbf{p}_s \in \mathcal{S}_i.
\end{equation}
Computing gradients of $g_{coll}$ for all $\mathbf{p}_s$ is computationally expensive. To accelerate the process, for each link \(\mathcal{I}_j\), we first construct a sparse active collision set by combining bounding-box filtering and conduct top-\(K\) smallest signed-distance selection:
\begin{equation}
    \mathcal{A}_{i,\mathcal{I}_j}^{K}
    =
    \operatorname*{TopK}_{\mathbf{p}_s\in
    \left\{
        \mathbf{p}_s\in\mathcal{S}_i
        \mid
        {}^{L_j}\mathbf{p}_{L_js}\in\mathcal{B}_{L_j}
    \right\}}
    \left(
        - d_{coll}^{\mathcal{I}_j}(\mathbf{p}_s, \mathbf{q}_i)
    \right).
    \label{eq:active_collision_set}
\end{equation}
Only samples in \(\mathcal{A}_{i,\mathcal{I}_j}^{K}\) are used for evaluating collision gradients $\nabla_{\mathbf{q}_i} g_{coll}$. It worth mentioning that after optimization, collision feasibility is evaluated using the full sampled query set (i.e., the operation of Line 14 in Algorithm~\ref{alg:optimization}). In our pipeline, the self-collision between robot links is handled in the same manner by treating sampled points on one robot link as query points and evaluating them against the SDF of another target link after coordinate transformation.

Note that the collision query requires transforming each query point \(\mathbf{p}_s\) into \{LCS\}$_j$ as \({}^{L_j}\mathbf{p}_{L_js}\). Thus, the gradient of $g_{coll}$ with respect to $\mathbf{q}_i$ can be derived using the chain rule.
\begin{equation}
    \begin{aligned}
    & \nabla_{\mathbf{q}_i} g_{coll}^{\mathcal{I}_j,\mathbf{p}_s}(\mathbf{q}_i) = -\left[\nabla_{{}^{L_j}\mathbf{p}_{L_js}} d^{\mathcal{I}_j}_{coll}(\mathbf{p}_s, \mathbf{q}_i)\right]^\top \frac{\partial {}^{L_j}\mathbf{p}_{L_js}}{\partial \mathbf{q}_i} \\
    &=\nabla \phi^{\mathcal{I}_j}_{SDF}({}^{L_j}\mathbf{p}_{L_js})^\top    {}^{L_j}\mathbf{R}_{X}
    \left(
        \mathbf{J}_{\mathbf{v},X,L_j}
        -
        \mathbf{S}({}^{X}\mathbf{p}_{Xs})
        \mathbf{J}_{\boldsymbol{\omega},X,L_j}
    \right).     
    \label{eq:gradient_sdf}
    \end{aligned}
\end{equation}
Here, the point-Jacobian term $\partial{}^{L_j}\mathbf{p}_{L_js}/\partial\mathbf{q}_i$ is acquired using differential kinematics:
\begin{equation}
\begin{aligned}
        {}^{L_j}\dot{\mathbf{p}}_{L_js}& = {}^{L_j}\dot{\mathbf{R}}_X{}^X\mathbf{p}_{Xs}-{}^{L_j}\mathbf{R}_X{}^X\dot{\mathbf{p}}_{XL_j} \\
        & = -{}^{L_j}\mathbf{R}_X\mathbf{S}({}^X\boldsymbol{\omega}_{L_j}){}^X\mathbf{p}_{Xs}-{}^{L_j}\mathbf{R}_X{}^X\dot{\mathbf{p}}_{XL_j} \\
        &=-{}^{L_j}\mathbf{R}_X(\mathbf{J}_{\mathbf{v},X,L_j}-\mathbf{S}({}^X\mathbf{p}_{Xs})\mathbf{J}_{\boldsymbol{\omega},X,L_j})\dot{\mathbf{q}}_i.
\end{aligned}
\end{equation}
\(X\in\{W,\,O,\,L_k\}\) specifies the frame in which 
\(\mathbf{p}_s\) is fixed. These choices correspond, respectively, to collision 
checking with obstacles, the evolving $\mathcal{O}_i$, and robot links, including 
inter-robot and self-collision cases. The matrices 
\(\mathbf{J}_{\mathbf{v},X,L_j}\) and 
\(\mathbf{J}_{\boldsymbol{\omega},X,L_j}\) denote the linear and angular components 
of the geometric Jacobian \(\mathbf{J}_{X,L_j}\), which maps the joint velocities 
to the twist of \{LCS\}$_j$ relative to frame \(X\), expressed in \(X\). 

When \(X=W\), \(\mathbf{J}_{W,L_j}\) reduces to the standard geometric Jacobian of 
link \(L_j\) with respect to the fixed base frame. When \(X=O\) or 
\(X=L_k\), the reference frame is itself moving. Hence, the relevant 
quantity is the relative motion of link \(L_j\) with respect to this dynamic frame. 
The corresponding Jacobian is therefore computed using the relative-Jacobian 
formulation introduced in Sec.~\ref{subsec:Relative Kinematics}.

%% file: tex/optimization.tex
\section{Effective Optimization by Manifold-guided Gradient Projection}
\label{sec:Optimization}

Given the differentiability of both the kinematic linkage and the collision model described above, this section details the proposed solver for the MAAM trajectory optimization. 

We first formulate the motion smoothness term (Eq.~\ref{eq:motionSmoothness}) and all inequality constraints (Eq.~\ref{eq:oriConstraint} - Eq.~\ref{eq:dynLimits}) as differentiable objectives. 
The deposition-path equality constraint (Eq.~\ref{eq:pathConstraint}) is enforced through iterative projection, which keeps each waypoint on the prescribed lower-dimensional path manifold with high numerical accuracy, followed by a manifold-constrained iterative solver with gradient projection. This design allows the optimizer to improve trajectory smoothness while maintaining path feasibility and penalizing constraint violations. 
    

\subsection{Loss Construction}

For a discretized joint trajectory \(\mathcal{Q}\), the inequality constraints are incorporated into the following penalized objective:
\begin{equation}
    \min_{\mathcal{Q}}
    \quad
    \mathcal{L}(\mathcal{Q})
    = \mathcal{L}_{obj}+\mathcal{L}_{ori}+\mathcal{L}_{coll}+\mathcal{L}_{joint} + \mathcal{L}_{dyn}
    \label{eq:penalizedObjective}
\end{equation}
\begin{equation}
    \begin{aligned}
    & \mathcal{L}_{obj}
    = \sum_{i=1}^N \Phi_i(\mathbf{v}_i,\mathbf{a}_i,\mathbf{j}_i), \\
    & \mathcal{L}_{ori}
    = w_{ori}\sum_{i=1}^N
    \left[
        \zeta(g_{ori}(\mathbf{q}_i),\delta_{ori})
        + \zeta(g_{grav}(\mathbf{q}_i),\delta_{grav})
    \right], \\
    & \mathcal{L}_{coll}
    = w_{coll}\sum_{i=1}^N\sum_{j}\sum_{\mathbf{p}_s}
    \zeta\!\left(
        g^{\mathcal{I}_j,\mathbf{p}_s}_{coll}(\mathbf{q}_i),
        \delta_{coll}
    \right), \\
    & \mathcal{L}_{joint}
    = w_{joint}\sum_{i=1}^N\sum_{j=1}^n
    \zeta\!\left(
        |q_{i,j}-\bar{q}_j| - \Delta q_j,
        \delta_{joint}
    \right), \\
    & \mathcal{L}_{dyn}
    = w_{dyn}\sum_{i=1}^N\sum_{j=1}^n
    \sum_{\mathbf{x}\in\{\mathbf{v},\mathbf{a},\mathbf{j}\}}
    \zeta\!\left(
        |x_j|-x_{max,j},
        \delta_{\mathbf{x}}
    \right).
\end{aligned}
\end{equation}
where $\zeta(\cdot)$ denotes the softplus penalty with safety margin
\begin{equation}
    \zeta(y,\delta) = \frac{1}{\beta}\log\!\big(1 + e^{\beta (y+\delta)}\big),
\end{equation}
and $\bar{q}_j = (q_{max,j}+q_{min,j})/2, \Delta q_j=(q_{max,j}-q_{min,j})/2$ in $\mathcal{L}_{joint}$. The path-following loss is omitted because the position constraint is explicitly handled by projection in Sec.~\ref{subsec:constrained_gradient_projection}.
The gradient of $\mathcal{L}$ with respect to $\mathcal{Q}$ can be stacked 
point-wise as
$\nabla_{\mathcal{Q}}\mathcal{L} 
    = [\,\nabla_{\mathbf{q}_1}\mathcal{L},\;\ldots,\;\nabla_{\mathbf{q}_N}\mathcal{L}\,]    \label{eq:stack}
$. For a generic penalty term $\zeta(g(\mathbf{q}),\delta)$, its gradient admits the
closed-form expression
\begin{equation}
    \nabla_{\mathbf{q}}\zeta(g(\mathbf{q}),\delta)
    =
    \sigma\!\left(\beta(g(\mathbf{q})+\delta)\right)
    \nabla_{\mathbf{q}} g(\mathbf{q}),
    \label{eq:softplusGrad}
\end{equation}
where $\sigma(y)=1/(1+\exp(-y))$ is the sigmoid function. 
All penalty terms therefore have closed-form gradients by the chain rule. 
The orientation and collision losses are differentiated as given in Sec.~\ref{subsec:Relative Kinematics} and Sec.~\ref{subsec:Modeling of Collision}, while the objective and dynamic losses are differentiated through the finite-difference operators for $\mathbf{v}_i$, $\mathbf{a}_i$, $\mathbf{j}_i$. 
Thus, $\nabla_{\mathcal{Q}}\mathcal{L}$ is obtained by summing the gradients of all loss components and is used in the subsequent constrained gradient projection step.

\begin{figure}[t]
    \centering
    \includegraphics[width = 0.9\linewidth]{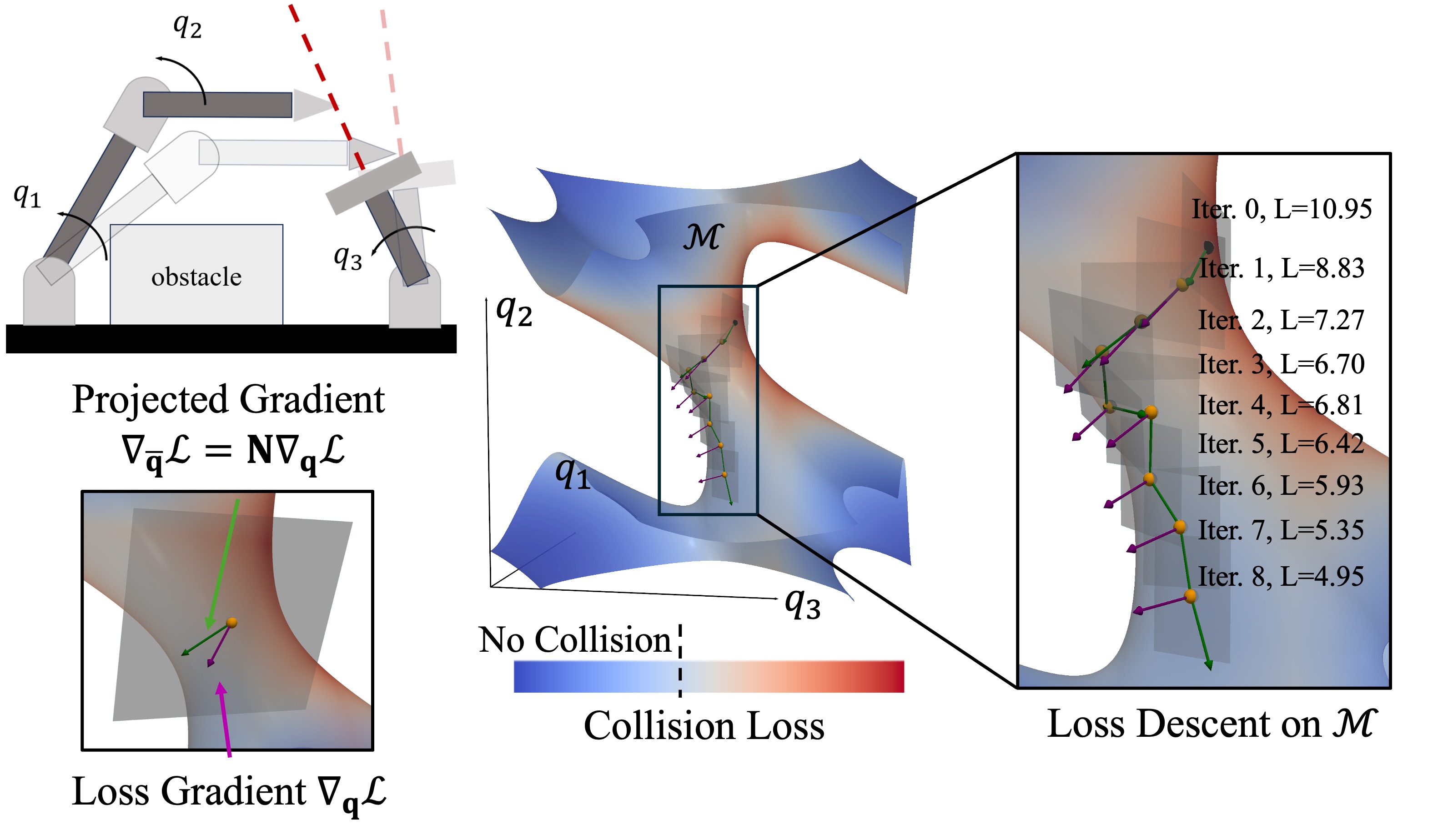}
\caption{
Projected gradient descent on the self-motion manifold $\mathcal{M}$ of a $2{+}1$-DoF dual-arm example. Gray planes denote the tangent space at corresponding configurations.
The descent gradient (purple vectors) is projected onto the tangent space to obtain a first-order task-preserving descent direction (green vectors), along which a configuration is updated to reduce collision penalties.}
    \label{fig:tangent_opt}
\end{figure}

\begin{algorithm}[t]
\small
\caption{Gradient-Projection Trajectory Optimization}
\label{alg:optimization}
\LinesNumbered

\KwIn{Initial trajectory \(\mathcal{Q}^{0}\), toolpath \(\{\mathbf{p}_i\}_{i=1}^{N}\), maximum iteration \(T_{\max}\), initial learning rate \(\alpha_{0}\), minimum learning rate \(\alpha_{\min}\), gradient tolerance \(\epsilon_g\), constraint tolerance \(\boldsymbol{\epsilon}_c\).}

\KwOut{Optimized trajectory \(\mathcal{Q}^{*}\).}

Set \(t=0\) and initialize the Adam optimizer state\;

\While{\(t<T_{\max}\)}{
    Project the current trajectory:
    \(\mathcal{Q}^{t}\leftarrow\Pi_{\mathcal{M}}(\mathcal{Q}^{t})\)\;

    \tcp{Constraint violation evaluation}
    \(\mathbf{c}^{t}\leftarrow\textsc{ConstraintCheck}(\mathcal{Q}^{t})\)\;

    Evaluate \(\mathcal{L}(\mathcal{Q}^{t})\) and its projected gradient
    \(\mathbf{g}_{\mathcal{L}}^{t}
    =
    \nabla_{\mathcal{Q}}\mathcal{L}(\mathcal{Q}^{t})\)\;

    \tcp{Feasible first-order stationarity}
    \If{\(\|\mathbf{g}_{\mathcal{L}}^{t}\|\le\epsilon_g\) \textbf{and} \(\mathbf{c}^{t}\le\epsilon_c\)}{
        \textbf{break}
    }

    \tcp{Cosine-decayed learning rate}
    \(\alpha_t=
        \alpha_{\min}
        +
        \frac{1}{2}(\alpha_0-\alpha_{\min})
        \left(
        1+\cos\left(\pi t/{T_{\max}}
        \right)\right);\)

    \tcp{Adam update}
    \(
        \mathcal{Q}^{t+1}
        \leftarrow
        \textsc{AdamUpdate}
        \left(
        \mathcal{Q}^{t},
        \mathbf{g}_{\mathcal{L}}^{t},
        \alpha_t
        \right);
    \)

    \(t=t+1\)\;
}

\(\mathcal{Q}^{*}=\Pi_{\mathcal{M}}(\mathcal{Q}^{t})\)\;

\tcp{Final constraint check}
\(\mathbf{c}^{*}\leftarrow\textsc{ConstraintCheck}(\mathcal{Q}^{*})\)\;

\Return{\(\mathcal{Q}^{*}, \mathbf{c}^{*}\)}

\end{algorithm}
\begin{figure*}[t]
    \centering
    \includegraphics[width=0.85\textwidth]{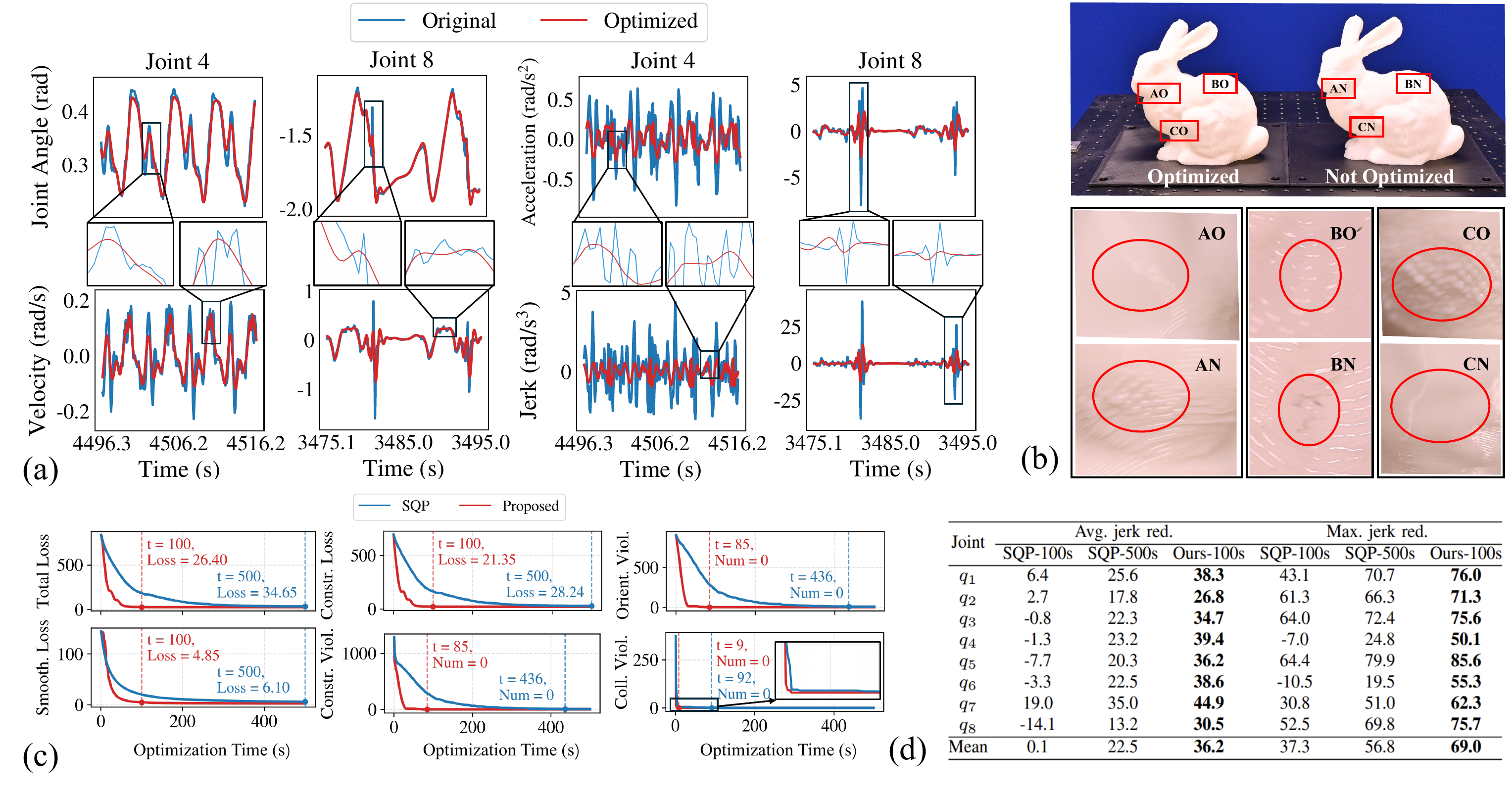}
    \caption{
    Computational and fabrication results on the support-free printing with the Stanford Bunny model. (a) Local joint-trajectory comparison before and after optimization. (b) Fabrication quality comparison with and without smoothness optimization. (c) Convergence analysis of different loss terms and constraint violations during optimization, showing great convergence for fast optimization over SQP-based methods~\cite{chen2025co}. (d) Per-joint jerk reduction after optimization.
    }
    \label{fig:bunny_joint_traj}
\end{figure*}

\subsection{Constrained Gradient Projection}
\label{subsec:constrained_gradient_projection}

Unlike inequality constraints, the deposition path constraint is enforced as a hard equality constraint. 
As illustrated in Fig.~\ref{fig:tangent_opt} for a low-DOFs example, at each waypoint \(\mathbf{p}_i\), the corresponding self-motion manifold can be constructed as
\begin{equation}
    \mathcal{M}_i
    =
    \left\{
    \mathbf{q}\in\mathbb{R}^{n}
    \mid
    \mathbf{FK}_{\mathbf{p}}(\mathbf{q})=\mathbf{p}_i
    \right\}.
    \label{eq:waypointManifold}
\end{equation}
Penalizing this equality constraint with a soft loss alone may allow small violations to accumulate, causing the trajectory to drift away from the prescribed toolpath (see~\cite{nocedal2006numerical} and detailed ablation study in Sec.~\ref{subsec:ablation}). We therefore enforce the deposition path constraint by projection rather than treating it merely as a penalty term. For each 
trajectory \(\mathcal{Q}\), every configuration is projected toward its corresponding manifold as
$    \bar{\mathcal{Q}}
    =
    \{\Pi_{\mathcal{M}_i}(\mathbf{q}_i)\}^N_{i=1}
    =
    \left\{ \bar{\mathbf{q}}_i \right\}_{i=1}^{N},
    \ 
    \bar{\mathbf{q}}_i \in \mathcal{M}_i,
    \label{eq:trajectoryProjection}$
, where the loss and its gradient (see zoom-in of Fig.~\ref{fig:tangent_opt}) are then evaluated on the projected trajectory 
\(\bar{\mathcal{Q}}\). This strategy keeps every waypoint being followed with high position accuracy needed for the additive manufacturing process.

To further reduce the numerical error introduced by the manifold-based projection, the operation \(\Pi_{\mathcal{M}_i}\) is implemented by an iterative Jacobian-based correction that drives each configuration onto  \(\mathbf{p}_i\). As illustrated in Algorithm~\ref{alg:optimization}, from 
\(\mathbf{q}_i^{0}=\mathbf{q}_i\) each waypoint is updated by
\begin{equation}
    \mathbf{q}_i^{k+1}
    =
    \mathbf{q}_i^{k}
    -
    \mathbf{J}_{\mathbf{v},i,k}^{\dagger}
    \left(
    \mathbf{FK}_{\mathbf{p}}(\mathbf{q}_i^{k})-\mathbf{p}_i
    \right),
    \label{eq:jacobianProjectionIteration}
\end{equation}
where \(\mathbf{J}_{\mathbf{v},i,k}=\partial \mathbf{FK}_{\mathbf{p}}(\mathbf{q}_i^{k})/\partial \mathbf{q}_i^{k}\) 
is the positional Jacobian evaluated at the current iterate \(\mathbf{q}_i^{k}\), and 
\(\mathbf{J}_{\mathbf{v},i,k}^{\dagger}\) denotes its damped 
pseudoinverse
    $\mathbf{J}_{\mathbf{v},i,k}^{\dagger}
    =
    \mathbf{J}_{\mathbf{v},i,k}^{\top}
    (
    \mathbf{J}_{\mathbf{v},i,k}\mathbf{J}_{\mathbf{v},i,k}^{\top}
    +
    \lambda \mathbf{I}
    )^{-1}$. In our implementation, a small fixed number 
$K_p$ of projection steps is used at each optimization iteration.
To back-propagate the loss gradient through the projection, we differentiate the 
iteration~\eqref{eq:jacobianProjectionIteration} with respect to the initial 
configuration. Denoting the position residual by 
\(\mathbf{e}_\mathbf{p}(\mathbf{q})=\mathbf{FK}_{\mathbf{p}}(\mathbf{q})-\mathbf{p}_i\), then
\begin{equation}
\begin{aligned}
    \nabla_{\mathbf{q}_i}\mathcal{L}&(\bar{\mathcal{Q}})
    =
    \left(\frac{\partial \bar{\mathbf{q}}_i}{\partial \mathbf{q}_i}\right)^{\!\top}
    \nabla_{\bar{\mathbf{q}}_i}\mathcal{L}
    =
    \prod_{k=K_p-1}^{0} \left(
    \frac{\partial \mathbf{q}_i^{k+1}}{\partial \mathbf{q}_i^{k}}
    \right)^{\!\top}
    \nabla_{\bar{\mathbf{q}}_i}\mathcal{L} \\
    = & 
    \prod_{k=K_p-1}^{0}
       \left(\mathbf{I}-\mathbf{J}_{\mathbf{v},i,k}^{\dagger}\frac{\partial \mathbf{e}_{\mathbf{p}}}{\partial \mathbf{q}_i^k}-\frac{\partial \mathbf{J}_{\mathbf{v},i,k}^{\dagger}}{\partial \mathbf{q}_i^k}\mathbf{e}_{\mathbf{p}}\right)^{\!\top}
    \nabla_{\bar{\mathbf{q}}_i}\mathcal{L}
    \\
    \approx &
    \prod_{k=K_p-1}^{0}
       {\underbrace{\left(\mathbf{I}-\mathbf{J}_{\mathbf{v},i,k}^{\dagger}\mathbf{J}_{\mathbf{v},i,k}\right)}_{\mathbf{N}_{\mathbf{v},i,k}}}^{\top} 
    \nabla_{\bar{\mathbf{q}}_i}\mathcal{L}. 
\end{aligned}
\end{equation}
The approximation holds when \(\mathbf{e}_{\mathbf{p}}\!\to\!\mathbf{0}\). Thus, for stable optimization, we first apply an unconstrained IK projection to satisfy the positional constraint and drive \(\mathbf{e}_{\mathbf{p}}\) close to zero before optimizing the remaining objectives and constraints in the projected space. $\mathbf{N}_{\mathbf{v},i,k}$ is actually the tangent-space projector. The full trajectory gradient \(\nabla_{\mathcal{Q}}\mathcal{L}(\bar{\mathcal{Q}})\) 
is finally assembled by stacking 
\(\{\nabla_{\mathbf{q}_i}\mathcal{L}(\bar{\mathcal{Q}})\}_{i=1}^{N}\).

With this gradient obtained, at each iteration the current trajectory is projected onto the path manifold, the closed-form gradient is evaluated, and an update step is taken using the Adam solver (see Line 10 of Algorithm~\ref{alg:optimization}). In our optimization scheme, a cosine-decayed learning rate is used for each iteration. All hyperparameter selections for the motion planning solver are listed in Table~\ref{tab:parameters}.



%% file: tex/experiment.tex
\section{Results and Discussion}
\label{sec:Experiment}

\begin{table}[t]
\scriptsize 
    \centering
    \caption{Common parameters used across all experiments.}
    \label{tab:parameters}
    \begin{tabular}{lcc}
        \hline
        Parameter & Value & Units\\
        \hline
         Smoothness weights $w_{\mathbf{v}}, w_{\mathbf{a}}, w_{\mathbf{j}}$ &  1, 2, 10 & -- \\
         Dynamic limits $\mathbf{v}_{\max}, \mathbf{a}_{\max}, \mathbf{j}_{\max}$ & 1, 5, 30 & rad/s,  rad/s$^2$, rad/s$^3$ \\
         Joint weight matrix $\mathbf{W}$& $\mathbf{I}$ & -- \\
         Printing speed $v_{print}$&  20 & mm/s \\
         Orientation tolerance $\theta_{max}, \theta_{max}^g$ & $30, 45$ & $^\circ$ \\
         Collision gradient query number $K$ & 8 & -- \\
         $w_{ori},w^m_{ori},w_{coll},w_{joint},w_{dyn}$ & 1e6 & -- \\
         Softplus smoothing parameter
         $\beta$ & 1e3 & -- \\
         Safety margin
         $\delta_{ori},\delta_{grav},\delta_{joint}$ & 1e-1 & -- \\
         Dynamic safety margin $\delta_{\mathbf{v}}, \delta_{\mathbf{a}}, \delta_{\mathbf{j}}$ & 1e-1, 5e-1, 3 & rad/s, rad/s$^2$, rad/s$^3$ \\
         Collision safety margin
         $\delta_{coll}$ & 3e-3 & m \\
         Damping parameter $\lambda $ & 1e-4 & -- \\
         Maximum projection steps $K_p$ & 2 & -- \\
         Optimizer $T_{max}, \alpha_0, \alpha_{\min}$ & 500, 5e-3, 2.5e-4 & -- \\
        \hline
    \end{tabular}
\end{table}

\begin{table*}[t]
\centering
\setlength{\aboverulesep}{0pt}
\setlength{\belowrulesep}{0pt}
\caption{Summary of experimental results and performance metrics. 
Vel., Acc., and Jerk denote the percentage reduction of the maximum 
joint velocity, acceleration, and jerk relative to the original
trajectory. $T_{coll}$, $T_{ori}$, and $T_{dyn}$ are the elimination times (s) at which all 
collision, orientation, and dynamic-limit violations are resolved during 
optimization, respectively. $P_e$ is the mean position error (mm) 
of the final trajectory. $T_{opt}$ is the total optimization time for the case.}
\label{tab:benchmark_summary}
\setlength{\tabcolsep}{4pt}
\renewcommand{\arraystretch}{1.0}
\begin{tabular}{llccc|cccccccc}
\toprule
\multirow{2}{*}{Category} 
& \multirow{2}{*}{Case} 
& \multirow{2}{*}{Size(mm$^3$) }
& \multirow{2}{*}{Fig.} 
& \multirow{2}{*}{Pts.} 
& \multicolumn{8}{c}{Metric} \\
\cmidrule(lr){6-13}
& & & & 
& Vel.\(\downarrow\)(\%) 
& Acc.\(\downarrow\)(\%) 
& Jerk\(\downarrow\)(\%) 
& \(T_{coll}\)(s) 
& \(T_{ori}\)(s) 
& \(T_{dyn}\)(s) 
& \(P_e\)(mm) 
& \(T_{opt}\)(s) \\
\midrule
\multirow{3}{*}{\shortstack{Support-free\\spatial printing}}
& Bunny 
& 168 $\times$ 130 $\times$ 182
& ~\ref{fig:setup_diagram}(a)
& 52,771 
& 12.7 
& 38.2 
& 75.2 
& 8.8 
& 84.6 
& 1.2
& 6.2e-3
& 206.8 \\

& Bone 
& 138 $\times$ 76 $\times$ 200
& ~\ref{fig:Teaser}(c)
& 36,904 
& 11.7 
& 21.5 
& 66.7 
& 3.5 
& 18.1 
& 5.7
& 5.1e-3 
& 184.4 \\

& TO-Bracket 
& 182 $\times$ 159 $\times$ 194
& ~\ref{fig:connector_result}
& 89,105 
& 8.8 
& 17.6 
& 61.3 
& 73.4 
& 92.5 
& 44.3
& 5.7e-3 
& 244.8\\
\midrule
\multirow{3}{*}{\shortstack{Conformal\\surface printing}}
& Bone 
& 138 $\times$ 76 $\times$ 200
& ~\ref{fig:Teaser}(c)
& 5,469 
& 2.1 
& 10.1 
& 58.8 
& 152.6 
& 9.4 
& 57.1
& 4.2e-3 
& 156.1 \\

& Dome 
& 200 $\times$ 100 $\times$ 32
& ~\ref{fig:dome}
& 59,323 
& 18.8 
& 41.2 
& 77.6 
& 1.1 
& 19.2 
& 2.2
& 4.5e-3 
& 197.2 \\

& Socket 
& 144 $\times$ 141 $\times$ 228
& ~\ref{fig:socket_result}
& 6,689 
& 15.3 
& 31.2 
& 71.4 
& 2.4 
& 7.5 
& 21.1
& 5.3e-3 
& 135.1\\
\bottomrule
\end{tabular}
\end{table*}

This section presents both computational and fabrication results, followed by an ablation study and comparison with existing work. We implement the proposed trajectory optimization pipeline in Python and run it on Ubuntu 24.04 using an Intel Core i9-285K CPU, 64 GB of RAM, and an NVIDIA RTX 5080 GPU. For physical fabrication, experiments were conducted on an 8-DOF ABB system equipped with an IRB 1200 robot arm and an IRB 250 positioner. The Dyze Design Typhoon extruder with a 1.2 mm nozzle is used to print polylactic acid (PLA) material at $200~^{\circ}$C.


\subsection{Computational and Fabrication Results}

We evaluated the proposed pipeline on six spatial toolpaths (Table~\ref{tab:benchmark_summary}), three for support-free spatial printing and three for conformal surface printing. With the proposed framework, the trajectories are consistently smoothed while satisfying the collision-free constraints across all tested cases. Specifically, the maximum joint velocity, acceleration, and jerk are reduced by up to $18.8\%$, $41.2\%$, and $77.6\%$, respectively, compared with the original trajectories. Most importantly, with the manifold-constrained projection scheme, the final mean position errors are controlled less than $10~\mu m$ across all test cases with collision-free motion being guaranteed. This confirms that the proposed solver can improve motion smoothness and enforce multiple constraints without compromising the prescribed deposition-path accuracy.

\begin{figure}
\centering
\includegraphics[width=0.85\linewidth]{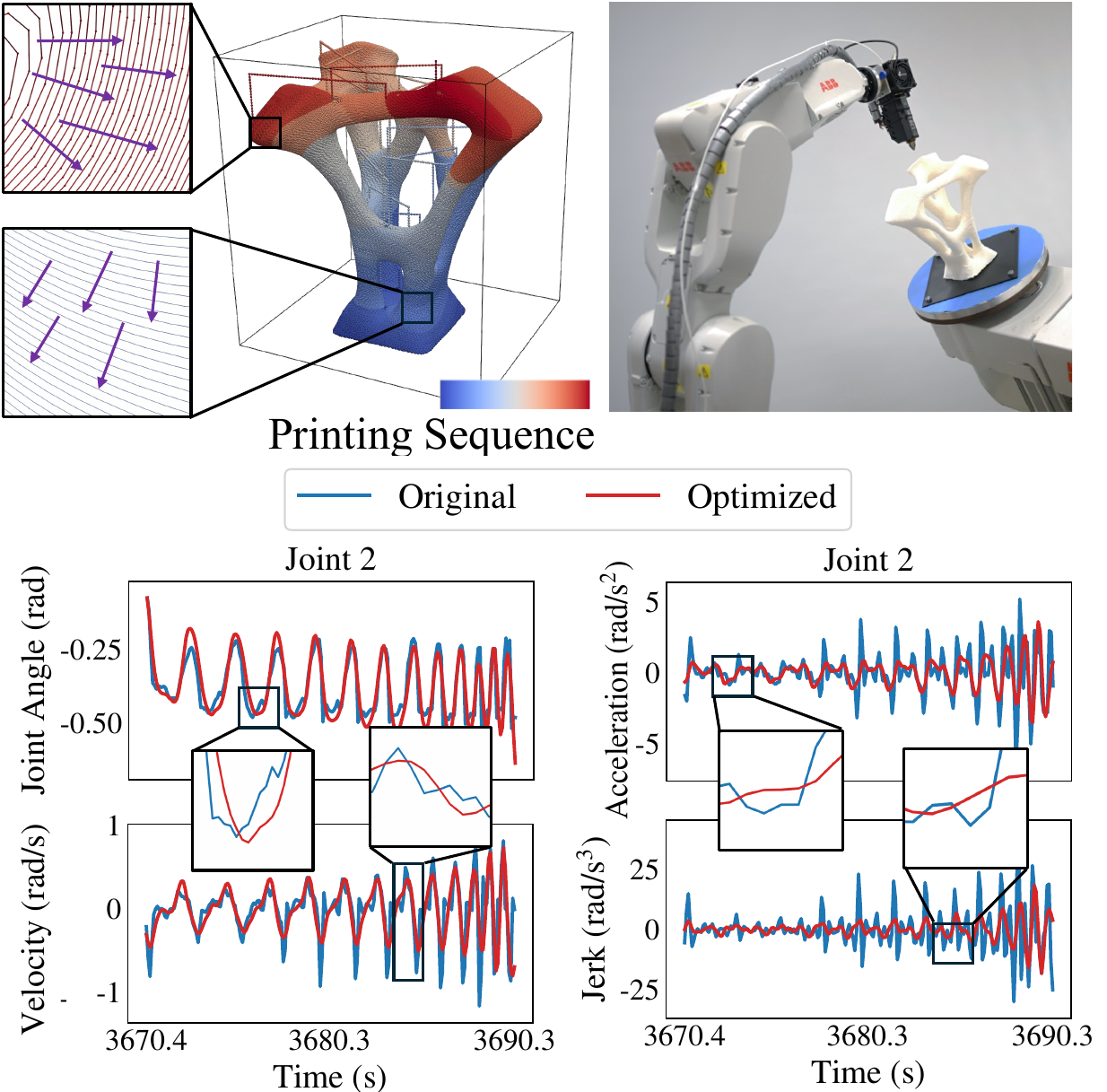}
\caption{Support-free spatial fabrication of a topology-optimized bracket and representative local comparison of joint trajectories before and after optimization.}
\label{fig:connector_result}
\end{figure}

\begin{figure}[t]
\centering
\includegraphics[width=1.0\linewidth]{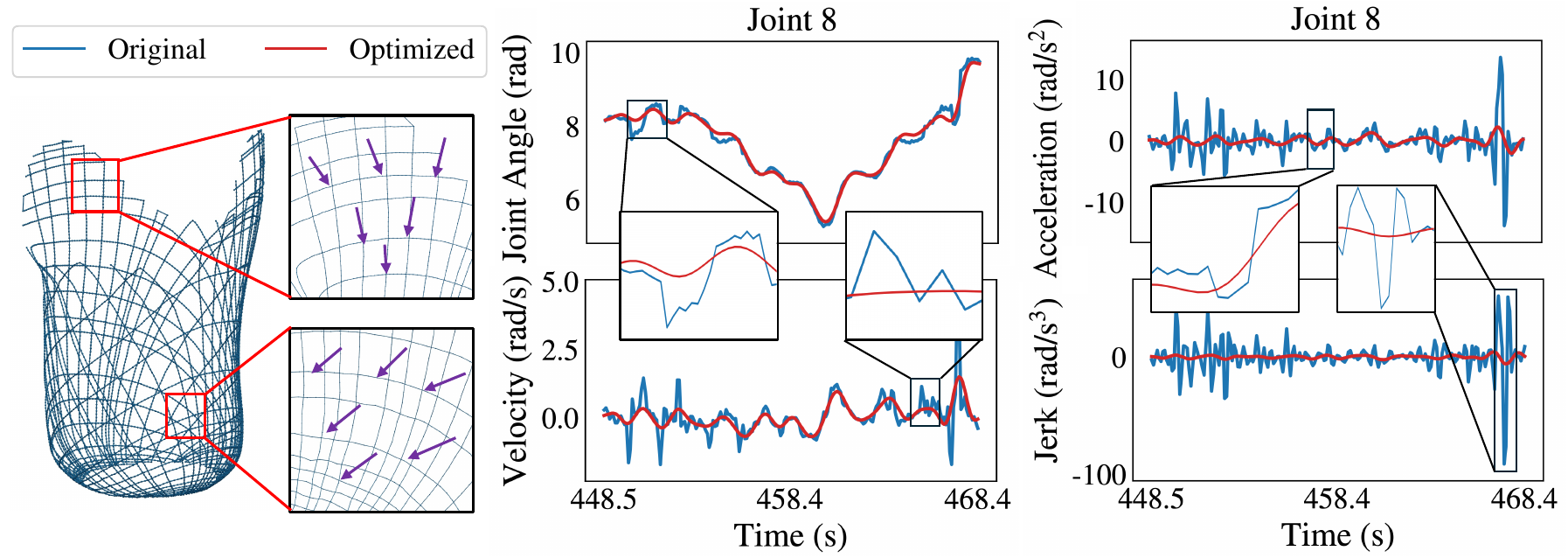}
\caption{Surface-conformal printing on a prosthetic socket and representative local comparison of joint trajectories before and after optimization.}
\label{fig:socket_result}
\end{figure}

The importance of motion smoothness is further verified through physical fabrication experiments - details can be found in the supplementary video. As shown in Fig.~\ref{fig:bunny_joint_traj}(a), for the Stanford bunny model, the optimized trajectory preserves the main joint-space motion trend while significantly reducing abrupt local variations. These improvements directly affect the physical printing quality. As shown in Fig.~\ref{fig:bunny_joint_traj}(b), the bunny fabricated with the optimized trajectory exhibits smoother material deposition and fewer visible surface artifacts. In contrast, the non-optimized trajectory produces noticeable defects, including local ripples, uneven bead accumulation, and rough surface textures (see zoom-in figure for details).

Additional results on the support-free spatial fabrication of a topology-optimized bracket and the surface-conformal printing on a prosthetic socket further demonstrate the performance and scalability of the proposed optimization pipeline. As shown in Fig.~\ref{fig:connector_result}, the proposed method successfully generates a feasible multi-axis trajectory for the TopOpt-bracket, whose long-horizon toolpath contains more than $89k$ waypoints and involves complex spatial deposition directions. After optimization, the local joint profiles preserve the overall motion trend while suppressing high-frequency variations, leading to smoother robot execution and successful physical fabrication without the need for supporting structures. Quantitatively, as shown in Table~\ref{tab:add_model_additional_results}, our method achieves a maximum jerk reduction of $61.3\%$ for the topology-optimized bracket and $71.4\%$ for the prosthetic socket, respectively. 

\begin{figure}[t]
\centering
\includegraphics[width=0.95\linewidth]{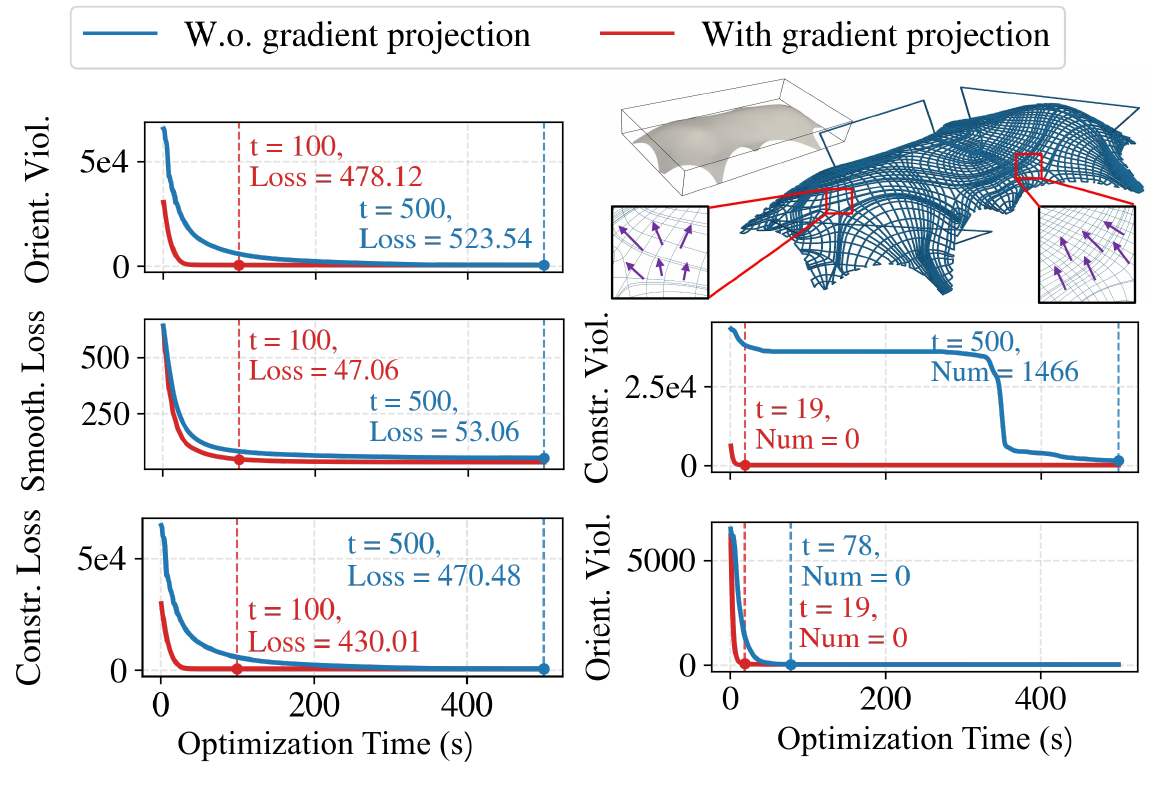}
\caption{Optimization process comparison between the proposed method with and without gradient projection in surface-conformal printing on a dome.}
\label{fig:dome}
\end{figure}

The computational results also demonstrate fast convergence and strong scalability with the proposed differentiable relative kinematics and collision model. As illustrated in Fig.~\ref{fig:bunny_joint_traj}(c), Fig.~\ref{fig:dome}, and Fig.~\ref{fig:bone_result}(a), collision-free, orientation, and dynamic-limit violations are resolved quickly even for challenging cases with severe initial violations (e.g., conformal printing on the bone model with large concave regions). For the bunny model, collision violations are eliminated within 10\,s for a path of more than 50k waypoints. Note that these times mark only when feasibility is first reached, while smoothness keeps improving afterward, as discussed in the next paragraph.

\begin{table}[t]
\scriptsize
\centering
\setlength{\aboverulesep}{0.2pt}
\setlength{\belowrulesep}{0.2pt}
\renewcommand{\arraystretch}{1.0}
\caption{Quantitative results for T.O. bracket and socket.}
\label{tab:add_model_additional_results}

\begin{subtable}{\linewidth}
\centering
\caption{Comparison on Jerk reduction.}
\label{tab:add_model_jerk_reduction_comparison}
\setlength{\tabcolsep}{3.0pt}
\begin{tabular}{lcccccc}
\toprule
\multirow{2}{*}{Case} 
& \multicolumn{3}{c}{Avg. jerk reduction (\%)} 
& \multicolumn{3}{c}{Max. jerk reduction (\%)} \\
\cmidrule(lr){2-4} \cmidrule(lr){5-7}
& SQP-100s & SQP-500s & Ours-100s 
& SQP-100s & SQP-500s & Ours-100s  \\
\midrule
TO-Bracket
& -1.2 & 25.5 & \textbf{27.2} 
& 27.8 & 58.9 & \textbf{61.3} \\
Socket 
& 2.7 & 49.6 & \textbf{28.4} 
& 33.1 & 62.0 & \textbf{71.4} \\
\bottomrule
\end{tabular}
\end{subtable}

\vspace{0.3em}

\begin{subtable}{\linewidth}
\centering
\caption{Constraint elimination time.}
\label{tab:add_model_constraint_elimination_time}
\setlength{\tabcolsep}{5pt}
\newcommand{\metriccell}[1]{\makebox[0.85cm][c]{#1}}
\begin{tabular}{llccc}
\toprule
\multirow{2}{*}{Case} 
& \multirow{2}{*}{Method} 
& \multicolumn{3}{c}{Constraint elimination time (s)} \\
\cmidrule(lr){3-5}
& 
& \metriccell{Ori.} 
& \metriccell{Coll.} 
& \metriccell{Dyn.} \\
\midrule
\multirow{2}{*}{TO-Bracket}
& SQP-based solver 
& \metriccell{281.1} 
& \metriccell{357.1} 
& \metriccell{101.5} \\
& This work 
& \metriccell{\textbf{73.4}} 
& \metriccell{\textbf{92.5}} 
& \metriccell{\textbf{44.3}} \\
\midrule
\multirow{2}{*}{Prosthetic socket}
& SQP-based solver 
& \metriccell{23.4} 
& \metriccell{82.7} 
& \metriccell{80.8} \\
& This work 
& \metriccell{\textbf{2.4}} 
& \metriccell{\textbf{7.5}} 
& \metriccell{\textbf{21.1}} \\
\bottomrule
\end{tabular}
\end{subtable}

\end{table}

\subsection{Comparison with Existing Method}

To further prove the scalability and effectiveness of the proposed trajectory optimization framework, we compared it with existing work using a direct SQP-based optimizer, which solves the constrained trajectory optimization by sequentially approximating it as quadratic subproblems with linearized constraints~\cite{chen2025co}.
Take the Stanford bunny model as an example (See Fig~\ref{fig:bunny_joint_traj}(c)), our method generally achieves lower losses in the same computation time. Our method also removes orientation violations by $t=85\,\mathrm{s}$, compared with $t=436\,\mathrm{s}$ for SQP-based solution ($5.1\times$ faster), and eliminates collision violations by $t=9\,\mathrm{s}$, compared with $t=92\,\mathrm{s}$ ($10.2\times$  faster). After these constraints converge, our method also shows great ability to ensure smooth motion at $t=100\,\mathrm{s}$, our result reduces the smoothness, and constraint losses to 4.85, and 21.35, respectively, while the SQP-based optimizer still has higher losses of 6.10, and 28.24 even running the optimization for $t=500\,\mathrm{s}$.  These results demonstrate higher convergence efficiency and improved satisfaction of the evaluated constraints.

\begin{figure}[t]
\centering
\includegraphics[width=0.95\linewidth]{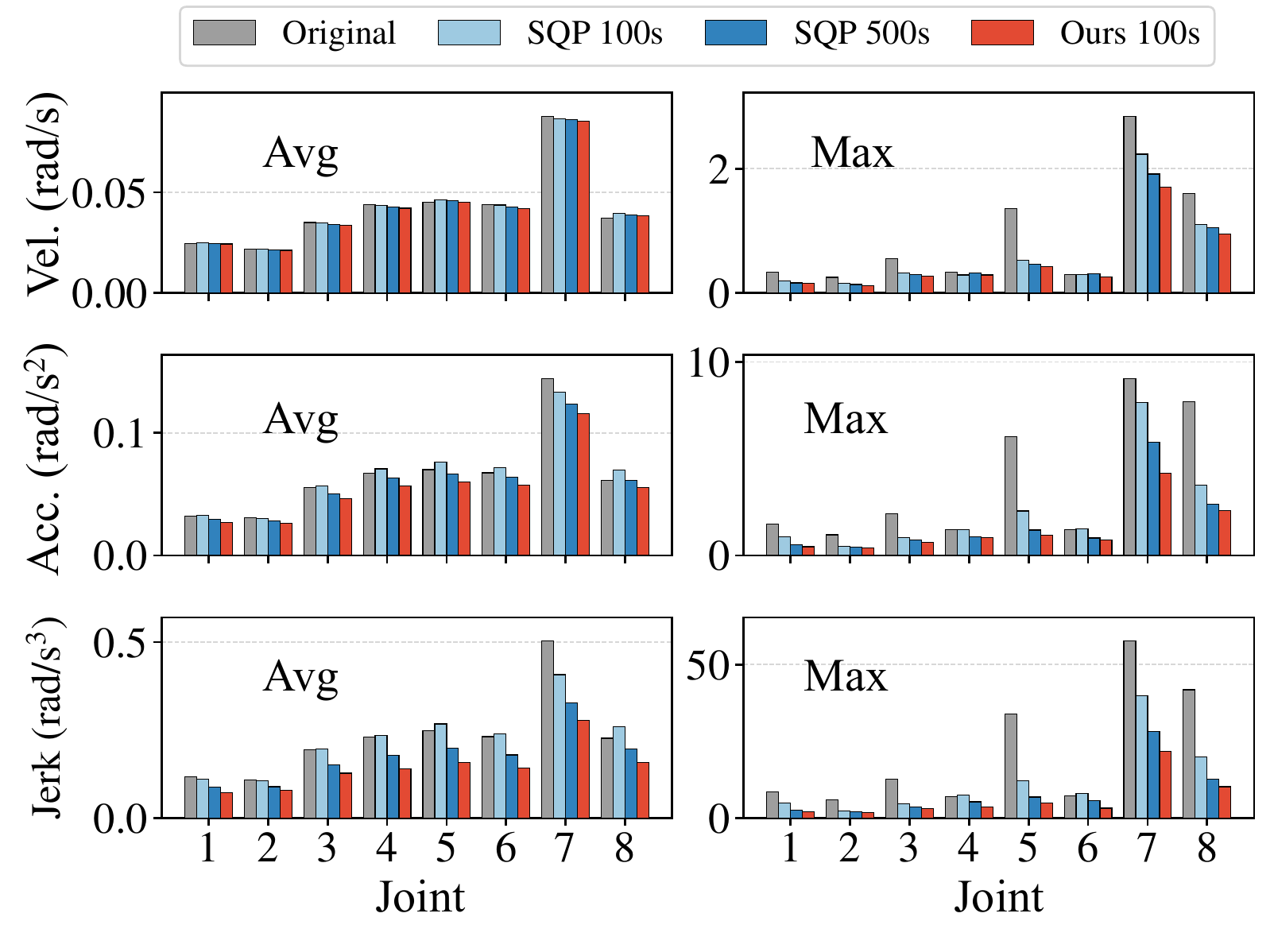}
\caption{Comparison of trajectory smoothness metrics among the support-free trajectory for the TO-Bracket model. The left column shows the average velocity, acceleration, and jerk of each joint, while the right column shows the corresponding maximum values.}
\label{fig:smoothness_comparison}
\end{figure}

We also quantitatively demonstrate the comparison on the toolpath with the TO-Bracket and Socket model. As demonstrated in Table~\ref{tab:add_model_additional_results}, the proposed method substantially improves high-order trajectory smoothness. Compared with the original trajectory, it reduces the average jerk of all joints by $26.8\%-44.9\%$ and the maximum jerk by more than $60\%$ within 100s running time. With the same computation time, the SQP-based time can even get worse motion smoothness (e.g., $1.2\%$ increase for the TO-Bracket model). Additionally, by looking at jerk reduction for individual axes as shown in Fig.~\ref{fig:smoothness_comparison}, the maximum jerk of $q_5$, $q_7$, and $q_8$ decreases from 39.8 to 5.7, from 61.2 to 25.1, and from 43.8 to 9.9, respectively. Moreover, the proposed method at 100\,s achieves lower average and maximum jerk than SQP-based method at 500\,s for all joints. This advantage stems from our fully parallelized formulation, which evaluates loss and gradients over all waypoints simultaneously and thus runs far more steps than SQP within the same time budget.

\subsection{Ablation Study on Manifold-based Gradient Projection}
\label{subsec:ablation}

\begin{figure}[t]
\centering
\includegraphics[width=1.0\linewidth]{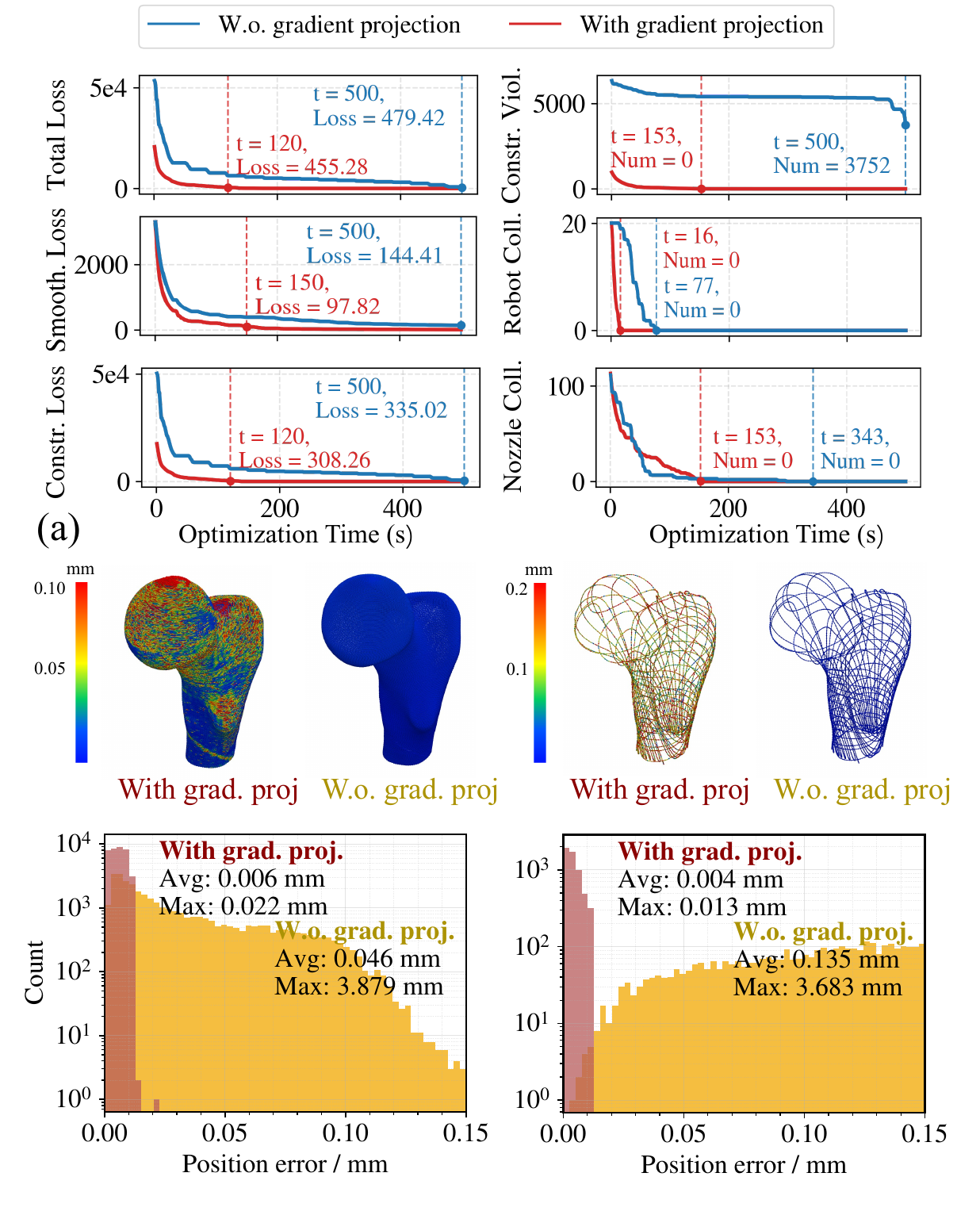}
\caption{(a) Optimization process comparison and (b) Position errors comparison between the proposed method with and without gradient projection in the two-phase MAAM task of a bone.}
\label{fig:bone_result}
\end{figure}

To evaluate the effect of the proposed gradient-projection strategy, we perform an ablation study by replacing the hard waypoint-wise deposition constraint with a soft penalty formulation~\cite{kang2020torm}. In the ablated variant, the manifold projection is removed, and the deposition-position error is directly included as an additional loss term. We compare the resulting convergence behavior and positional drift against the proposed method on representative MAAM toolpaths.


Take the dome model as an example (See Fig~\ref{fig:dome}), our method reduces the total, smoothness, and constraint losses to $478.12$, $47.06$, and $430.01$ at $t=100\,\mathrm{s}$, whereas the ablated version still exhibits higher losses of $523.54$, 
$53.06$, and $470.48$ even after $t=500\,\mathrm{s}$. The gradient projection 
strategy also eliminates the evaluated constraints more rapidly. As reported in Table~\ref{tab:ablation_comparison}, orientation, 
collision, and motion violations are removed at $t=19.2\,\mathrm{s}$, 
$1.1\,\mathrm{s}$, and $2.2\,\mathrm{s}$, compared with $81.7\,\mathrm{s}$, 
$11.8\,\mathrm{s}$, and $17.9\,\mathrm{s}$ for the ablated version 
($4.3\times$, $10.7\times$, and $8.1\times$ faster). A similar trend holds 
for the bone case, where our method removes robot and nozzle collision 
violations by $t=16\,\mathrm{s}$ and $t=153\,\mathrm{s}$, versus 
$t=77\,\mathrm{s}$ and $t=343\,\mathrm{s}$ without gradient projection 
($4.8\times$ and $2.2\times$ faster).

More importantly, gradient projection preserves positional accuracy throughout the optimization. As shown in Fig.~\ref{fig:bone_result} and Table~\ref{tab:ablation_comparison}, the maximum position error of the ablated version reaches $3.7$ - $3.9\,\mathrm{mm}$ across all tested cases, which is unacceptable for high-precision additve manufacturing task. As a comparsion, our method keeps the error within $1.3$ - $2.2\times10^{-2}\,\mathrm{mm}$ ($0.005\%$ of the model size), which showcase a two orders of precision improvement with manifold-based gradient projection scheme. 
These quantitative results support our claim that treating deposition accuracy as a soft penalty is insufficient for redundant robotic MAAM trajectory optimization. A penalty-based formulation cannot strictly enforce deposition accuracy and may trade it off against other objectives (e.g., motion smoothness and collision avoidance). By contrast, the proposed gradient-projection method preserves the deposition position and uses the remaining redundancy to improve feasibility and motion quality. 


\begin{table}
\scriptsize
\centering
\setlength{\aboverulesep}{0pt}
\setlength{\belowrulesep}{0pt}
\renewcommand{\arraystretch}{1.0}
\caption{Quantitative results for ablation study}
\label{tab:ablation_comparison}

\begin{subtable}{\linewidth}
\centering
\caption{Max jerk reduction and position error comparison.}
\label{tab:jerk_reduction_comparison}
\setlength{\tabcolsep}{6pt}
\begin{tabular}{lcccc}
\toprule
\multirow{2}{*}{Case} 
& \multicolumn{2}{c}{Max. jerk reduction (\%)} 
& \multicolumn{2}{c}{Max. pos. reduction (mm)} \\
\cmidrule(lr){2-3} \cmidrule(lr){4-5}
& W.o. & With
& W.o. & With \\
\midrule    
Dome 
& 43.9 & \textbf{77.6} & 3.7 & \textbf{1.5e-2} \\
Bone (Spatial)
& 36.7 & \textbf{66.7} & 3.9 & \textbf{2.2e-2} \\
Bone (Conformal)
& 31.3 & \textbf{58.8} & 3.7 & \textbf{1.3e-2} \\
\bottomrule
\end{tabular}
\end{subtable}

\vspace{0.3em}

\begin{subtable}{\linewidth}
\centering
\caption{Constraint elimination time.}
\label{tab:constraint_elimination_time}
\setlength{\tabcolsep}{5pt}
\newcommand{\metriccell}[1]{\makebox[1.05cm][c]{#1}}
\begin{tabular}{llccc}
\toprule
\multirow{2}{*}{Case} 
& \multirow{2}{*}{Method} 
& \multicolumn{3}{c}{Constraint elimination time (s)} \\
\cmidrule(lr){3-5}
& 
& \metriccell{Orientation} 
& \metriccell{Collision}
& \metriccell{Motion} \\
\midrule
\multirow{2}{*}{Dome}
& W.o. 
& \metriccell{81.7} 
& \metriccell{11.8} 
& \metriccell{17.9} \\
& With
& \metriccell{\textbf{19.2}} 
& \metriccell{1.1} 
& \metriccell{2.2} \\
\midrule
\multirow{2}{*}{Bone (Spatial)}
& W.o. 
& \metriccell{95.1} 
& \metriccell{22.0} 
& \metriccell{36.4} \\
& With 
& \metriccell{\textbf{18.1}} 
& \metriccell{3.5} 
& \metriccell{\textbf{5.7}} \\
\midrule
\multirow{2}{*}{Bone (Conformal)}
& W.o.
& \metriccell{26.5} 
& \metriccell{342.7} 
& \metriccell{160.9} \\
& With
& \metriccell{\textbf{9.4}} 
& \metriccell{\textbf{152.6}} 
& \metriccell{\textbf{57.1}} \\
\bottomrule
\end{tabular}
\end{subtable}

\end{table}

\subsection{Discussion and Future Work}
With the presented results, the proposed method demonstrates strong convergence and scalability for trajectory optimization. However, its performance remains sensitive to initialization. Poor initial inputs for challenge tasks — such as those with severe joint-limit violations or deep collisions — may require substantially more iterations or remain infeasible within a fixed budget. We plan to address this through a hybrid strategy combining sampling-based search with local optimization. In addition, singularity avoidance is not explicitly constrained but only indirectly discouraged by the velocity and acceleration limits. We will incorporate a differentiable singularity-aware term based on the minimum singular value, condition number, or manipulability~\cite{chen2024toolpath} in future work.

The SDF-based collision model proposed here is evaluated at discrete waypoints and sampled geometry points. While effective and differentiable, its sampling density directly affects memory cost and runtime. In our implementation, the geometry of the collision model (i.e., nozzle and robot links) is sampled with an approximate spacing of $2 mm$ — well below the adopted safety margin — which reliably detects all collisions across the tested models. For larger-scale MAAM with redundant systems, representing the geometry as a neural SDF~\cite{qu2025inf} is a promising direction for better scalability and detection accuracy.


%% file: tex/conclusion.tex
\section{Conclusion}
\label{sec:Conclusion}


This paper presented a constrained gradient-projection framework for collision-aware trajectory optimization in redundant robotic multi-axis additive manufacturing. The proposed method enforces the prescribed deposition position as a hard waypoint-wise equality constraint by projecting gradient updates onto the tangent space of the self-motion manifold, and further develops a fabrication-sequence-aware differentiable collision model that accounts for the accumulated printed geometry. Experiments on an 8-DOF robotic MAAM platform verified its effectiveness and scalability on both support-free spatial and conformal surface printing. Compared with the SQP-based baseline, the proposed method achieved up to \(10.2\times\) faster convergence while maintaining substantially smaller position drift.

%% file: tex/appendix.tex
\appendices
\numberwithin{equation}{section}
\renewcommand{\theequation}{\thesection.\arabic{equation}}